%% file: main.tex
\documentclass[10pt,twocolumn,letterpaper]{article}

\usepackage[pagenumbers]{wacv} 

\usepackage{graphicx}
\usepackage{amsmath}
\usepackage{amssymb}
\usepackage{booktabs}
\usepackage{xcolor}
\usepackage{balance}

\definecolor{orange2}{HTML}{FF6600} 
\definecolor{blue2}{HTML}{0066FF}   
\definecolor{teal2}{HTML}{008080}

\usepackage[pagebackref,breaklinks,colorlinks]{hyperref}

\usepackage[capitalize]{cleveref}
\crefname{section}{Sec.}{Secs.}
\Crefname{section}{Section}{Sections}
\Crefname{table}{Table}{Tables}
\crefname{table}{Tab.}{Tabs.}

\def\wacvPaperID{2529} 
\def\confName{WACV}
\def\confYear{2025}

\begin{document}


\title{Psych-Occlusion: Using Visual Psychophysics for Aerial Detection\\ of Occluded Persons during Search and Rescue}

\author{Arturo Miguel Russell Bernal, Jane Cleland-Huang, Walter Scheirer\\
Computer Science and Engineering Department, University of Notre Dame, Indiana, USA\\
{\tt\small arussel8@nd.edu, janehuang@nd.edu, walter.scheirer@nd.edu}
}
\maketitle

\begin{abstract}
    \vspace{-0.2cm}
    The success of Emergency Response (ER) scenarios, such as search and rescue, is often dependent upon the prompt location of a lost or injured person.  With the increasing use of small Unmanned Aerial Systems (sUAS) as ``eyes in the sky" during ER scenarios, efficient detection of persons from aerial views plays a crucial role in achieving a successful mission outcome. Fatigue of human operators during prolonged ER missions, coupled with limited human resources, highlights the need for sUAS equipped with Computer Vision (CV) capabilities to aid in finding the person from aerial views. However, the performance of CV models onboard sUAS substantially degrades under real-life rigorous conditions of a typical ER scenario, where person search is hampered by occlusion and low target resolution.  To address these challenges, we extracted images from the NOMAD dataset and performed a crowdsource experiment to collect behavioural measurements when humans were asked to ``find the person in the picture". We exemplify the use of our behavioral dataset, Psych-ER, by using its human accuracy data to adapt the loss function of a detection model. We tested our loss adaptation on a RetinaNet model evaluated on NOMAD against increasing distance and occlusion, with our psychophysical loss adaptation showing improvements over the baseline at higher distances across different levels of occlusion, without degrading performance at closer distances.  To the best of our knowledge, our work is the first human-guided approach to address the location task of a detection model, while addressing real-world challenges of aerial search and rescue. All datasets and code can be found at: \href{https://github.com/ArtRuss/NOMAD}{https://github.com/ArtRuss/NOMAD}.
\end{abstract}
\vspace{-0.5cm}

\input{sec_introduction}
\input{sec_related_work}

\input{sec_psycho_data}

\input{sec_psycho_loss}
\input{sec_results}
\input{sec_futurework_conclusions}

\section{Acknowledgments}
\label{sec:Acknowledgements}
The work described in this paper was supported by the USA National Science Foundation under grant CNS-1931962. In addition, we thank all participants and annotators, as well as family and friends for their logistics support.

\balance
{\small
\bibliographystyle{ieee_fullname}
\bibliography{wacvbib}
}

\end{document}

%% file: sec_introduction.tex
\section{Introduction}
\label{sec:intro}

\begin{figure*}[t]
  \centering
   \includegraphics[trim=0cm 0.3cm 0cm 0.3cm,clip,width=\linewidth]{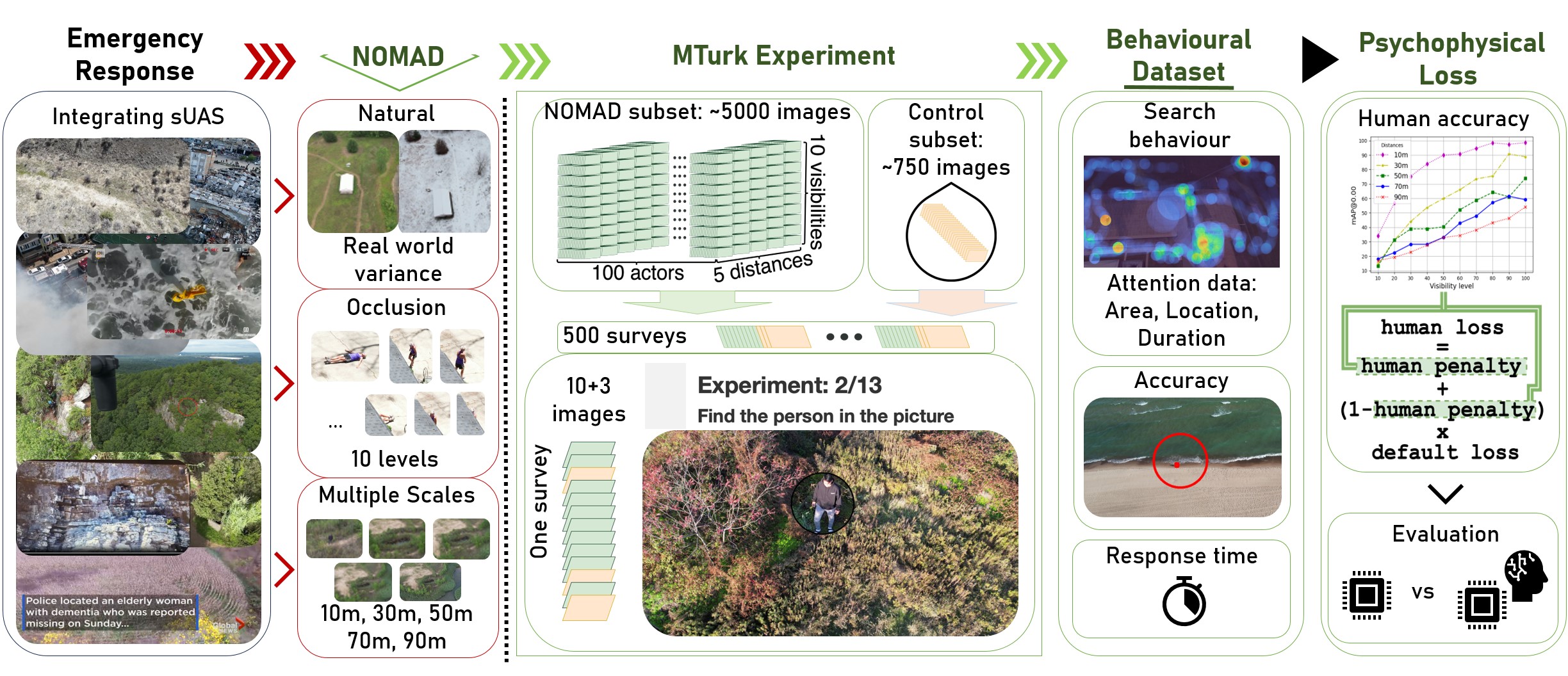}\vspace{-0.2cm}
   \caption{Development of Psych-ER, our behavioral dataset for Emergency Response (ER) aerial search, and its derived psychophysical loss.  Integration of sUAS into ER scenarios have aided first responders and rescued victims \cite{news_drowning, news_hikingAfrica, news_lostWoods, news_earthquake, news_girl, news_fire, news_teamwork, news_oldadult, news_firehomes} (first column). Onboard Computer Vision (CV) is a key component for the full integration of sUAS into ER missions; to address the inherent challenges of CV for ER scenarios we had previously published NOMAD\cite{russell2024nomad}, an ER dedicated dataset composed of 42,825 aerial images, filmed from five aerial distances and providing a label detailing the degree of occlusion of each person's bounding box (second column).  With the goal of improving CV performance for ER, we extracted a comprehensive subset of NOMAD and built an MTurk experiment to collect human behavioral measurements when facing the task ``Find the person in the picture" (third column). Our behavioral dataset Psych-ER contains (1) human search behavioral data for more than 5000 images collected through the participant's screen cursor, including location and area of attention, as well as duration at every attention location; (2) accuracy of their selection relative to NOMAD's ground truth bounding box; (3) response time for every image (fourth column). Finally, we used the human accuracy data to formulate a psychophysical loss, and evaluated our loss performance on the RetinaNet architecture.}
   \vspace{-0.5cm}
   \label{fig:main} 
\end{figure*}

Advances in technology have led to the increased use of small Unmanned Aerial Systems (sUAS) across a broad range of applications \cite{survey_civilApplications, survey_edgeAI_UAVs}. Emergency Response (ER) represents one of the fields that has been most benefited with the deployment of sUAS, where the sUAS versatility has aid ER missions through aerial surveillance, 3D mapping, architectural inspections, and delivery services, among other capabilities \cite{lyu2023unmanned}.  Mission success in ER scenarios, such as search and rescue, search for missing persons, and search for suspects, is most of the times defined on promptly finding the person in need, therefore ER demands rigorous and reliable performance of sUAS, as mission success can make the difference between life and death.  The detection task of a person in need, during manual deployment of sUAS in ER scenarios, usually includes a human pilot searching for a person through the video-stream of the sUAS' camera. However, factors such as fatigue of human operators during prolonged ER missions, coupled with limited human resources, highlights the need for sUAS equipped with Computer Vision (CV) capabilities to aid in finding the person from aerial views, allowing emergency responders to focus attention on mission level goals, while sUAS perform lower-level person detection tasks \cite{chi20-partnerships, hmt_cleland2024human, sophia_reliability_score}.

Despite the benefits of onboard CV capabilities, CV models are known to underperform when facing aerial views compared to ground data \cite{survey_aerial}, in addition, deploying CV onboard sUAS for ER scenarios brings non-trivial challenges \cite{challenges_tawfiq, challenges_murphy, russell2024nomad}, such as varying distance from the target, which modifies the ground field of view and the resolution of the target, and the highly prevalent problem of occlusion, which occurs when targets of aerial search are partially hidden from view.  For example, a drowning victim who is partially submerged in water, people buried in debris following an earthquake, hidden by smoke in a fire, laying behind trees and rocks in search and rescue missions, or intentional occlusion when a suspect is hiding from law-enforcement. Occlusion could also be caused by pose and image perspective, or introduced in far-distance aerial views due to glare, shades, blur, and low resolution.  

Human vision has shown significant robustness to occlusion \cite{walter_original, occlusionvshumans} and outperformed CV models on aerial detection of persons during search and rescue scenarios \cite{maruvsic2018region}. With previous research having successfully incorporated human perception into CV models in a variety of manners \cite{walter_original, dulay2024informing, adam_iris}, we addressed the challenge of person aerial detection under occlusion for ER scenarios through visual psychophysics.  Psychophysics is the systematic study of behavioral responses to physical changes in sensory stimuli \cite{gescheider2013psychophysics}, in our case, this corresponds to the systematic study of human behaviour when presented with aerial images of varying distance to ground and with varying level of occlusion of the person within the image. Images used for this purpose were extracted from NOMAD \cite{russell2024nomad}, an ER focused dataset with a distance label for every image and person occlusion labels for every bounding box. We built a crowdsource survey deployed through the MTurk platform to collect human behavioral measurements, where human participants were presented with one image at a time facing the task ``Find the person in the picture". Our behavioral dataset Psych-ER contains (1) human search behavioral data for more than 5000 images collected through the participant's screen cursor, including location and area of attention, as well as duration at every attention location; (2) accuracy of their selection relative to NOMAD's ground truth bounding box; (3) response time for every image.  We analyzed the psychophysical data and computed accuracy plots that were used to formulate an adaptation to the bounding box regression loss of a CV detection model.  To test our approach we used the NOMAD dataset, and trained a RetinaNet-R101-FPN model from the Detectron2 library \cite{detectron2modelzoo}, with and without our loss adaptation, evaluating our approach against different levels of distance and occlusion, with our psychophysical loss adaptation showing improvements over the baseline at higher distances across different levels of occlusion, without degrading performance at closer distances.

Our contributions are two-folded: (1) we provide to the community Psych-ER, a dataset of human behavioural measurements, containing more than 5000 aerial images from the NOMAD dataset, annotated with accuracy, response time and human search behavior for person aerial detection; (2) we exemplify the use of the behavioral dataset by formulating a human-guided bounding box regression loss. Previous human-guided approaches have mostly explored the CV task of classification, to the best of our knowledge, our work is the first human-guided approach to address the location task of a CV detection model, while addressing real-world challenges of aerial search and rescue, such as varying distance to the target and the prevalence of occlusion.

The remainder of this article is organized as follows. \Cref{sec:related_work} presents related work. \Cref{sec:psycho_data} describes Psych-ER's data collection process, including its data analysis. \Cref{sec:psycho_loss} describes the formulation of the detection loss. \Cref{sec:results} reports the performance results on a RetinaNet model across different levels of distance and occlusion. Finally,  \Cref{sec:future_work_conclusions} describes future work and conclusions.

%% file: sec_related_work.tex
\section{Related Work}
\label{sec:related_work}

\begin{figure}[t]
  \centering
   \includegraphics[width=\linewidth]{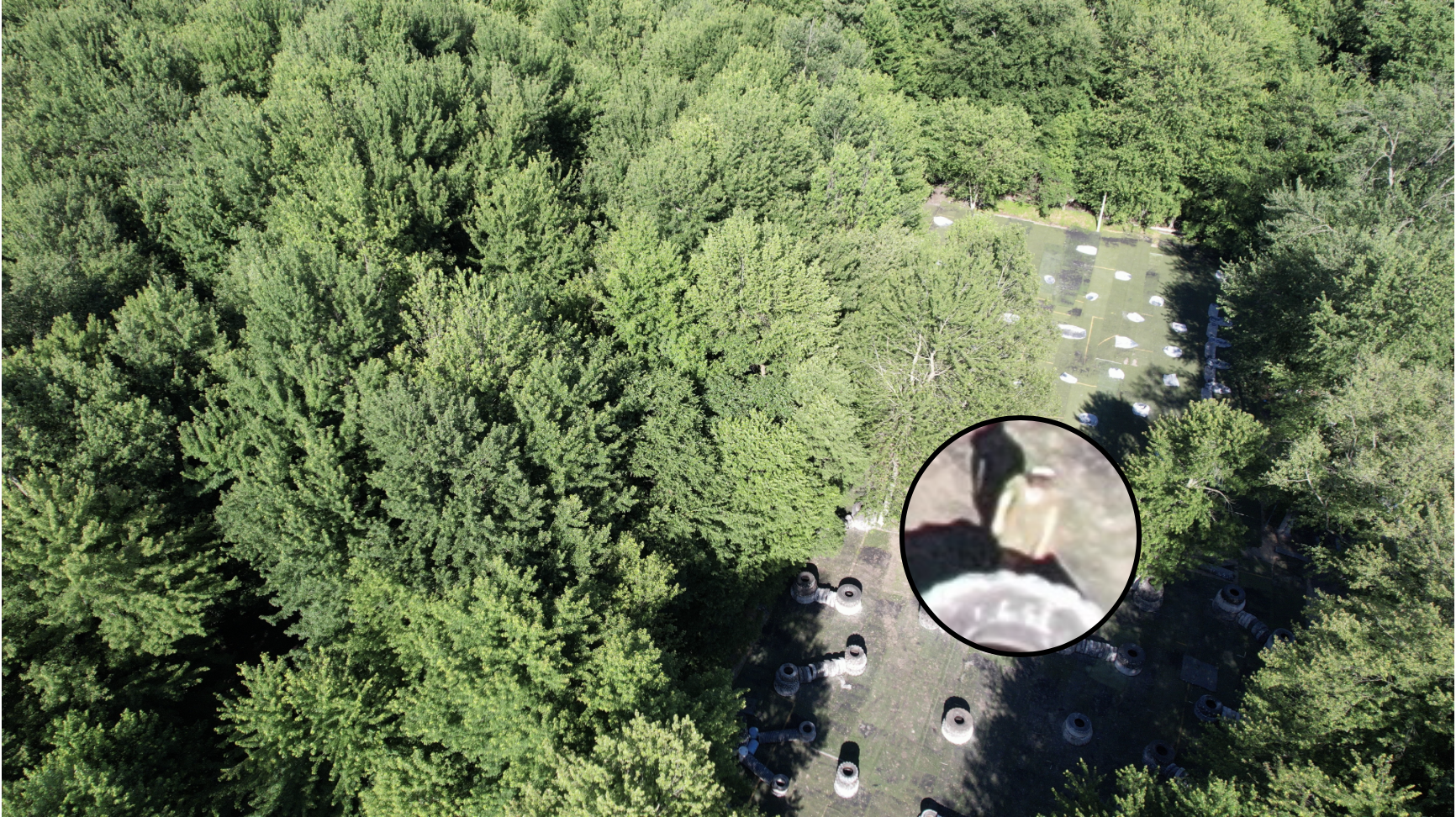}
   \caption{Sample interactive picture shown during the instructions of the MTurk "Find the person in the picture" survey, explaining the use of the provided zoom-magnifying-glass.}
   \vspace{-0.3cm}
   \label{fig:zoom}
\end{figure}

\subsection{Mobile Robotics for Emergency Response}
Previous literature has identified a number of challenges associated with integrating  mobile robotics into emergency response missions  \cite{challenges_tawfiq, challenges_murphy, challenges_walter, survey_civilApplications, survey_edgeAI_UAVs, nuclear_accident}.  Focusing on ground mobile robots, some researchers have developed specialized architectures \cite{ground_gasNose, ground_snake}, improved communication networks \cite{ground_network}, and addressed the challenge of emergency scenes' mapping \cite{ground_map_reinforcementLearning, ground_3dmap, ground_SMURF}, with others exploring the collaboration with aerial robots within these challenges \cite{aerial_ground_collaboration,  aerial_indoors, aerial_mapping}.  Wankm\"uller \etal \cite{benefit_usability_tests} performed usability studies that demonstrated the benefits of including aerial robots in emergency response, stressing that several improvements are required. Researchers working with aerial robots have explored self-adaptation against anomalies during missions \cite{islam2024adam}, as well as the integration of additional sensors, such as ground penetrating radars \cite{aerial_groundpenetratingradar}, or cellphone tracking capabilities for missing person search \cite{aerial_trackingSignal}. Additionally, several researchers have explored efficient collaborations between humans and sUAS at the intersection of software engineering and human computer interaction \cite{hci_murphy, chi20-partnerships, hmt_Cleland-HuangAV22, hmt_cleland2024human, chambers2024self, chambers2024hifuzz, droneresponse, granadeno2024environmentally}.

\subsection{Real-World Object Detection}
There are further challenges related to utilizing aerial CV for real-time emergency response \cite{survey_generic_object_detection, lowpowercv, object_detection_survey, survey_aerial}.  Real-time CV applications tend to lean towards the YOLO algorithm \cite{redmon2016yolo, yolov4, yolov7, yolov8_ultralytics} and its modifications \cite{driving_yolo, edge_yolo_wacv, pedestrian_yolo, object_afc_edge_yolo, ppyoloe}. Other methods explore attention for object detection \cite{zhao2024detrs, attention_object_detection}, deviate from the bounding box architecture \cite{law2018cornernet, zhou2019objects, duan2019centernet}, explore multimodal techniques \cite{pedestrian_multimodal}, as well as the Open World Object Detection modality \cite{bendale2015towards, prijatelj2022human, OWOD_CVPR_birth, } with its variations \cite{UCOWOD, OSODD}. The challenges of object detection for low resolution \cite{survey_low_resolution, multires_drones} and under occlusion \cite{survey_generic_occlusion, survey_pedestrian_occlusion, occlusion_oldsurvey} have also been studied.   In the field of emergency response, several aerial datasets have been created to aid CV models \cite{russell2024nomad, wisard, custom_rcnn_heridal, sard}, with architectural \cite{airsight, sophia_reliability_score} and algorithmic \cite{custom_rcnn_heridal, deep_cnn_sard} contributions.  To the best of our knowledge, no previous work has been done to address aerial detection of persons under search and rescue conditions, through visual psychophysics.

\subsection{Visual Psychophysics in Computer Vision}
Researchers have studied the performance of humans vs machines in the field of CV \cite{occlusionvshumans, moreira2019performance, trokielewicz2019perception}, proposing psychophysical frameworks to evaluate model's performance \cite{richardwebster2018visual, psyphy}, presenting human inspired techniques \cite{humanperception_objectdetection, occlusion_humanperception}, and exploring the idea of improving CV models by embedding human behavioral data into them, as inputs to the model \cite{adam_iris, adam_new, boyd2022human}, architectural modifications \cite{linsley2018learning, crum2023teaching, crum2023mentor}, or loss regularizers \cite{walter_original, walter_brain, sam_writting, huang2023measuring, dulay2024informing, justin_tl, boyd2023cyborg, piland2023model}.  The integration of human perception into models has been explored in a variety of CV applications, including face detection \cite{walter_original} and synthetic face classification \cite{boyd2023cyborg, piland2023model, crum2023teaching}, iris recognition and attack detection \cite{adam_iris, adam_new, boyd2022human}, object recognition \cite{walter_brain, linsley2018learning} and open set recognition \cite{huang2023measuring}, as well as character recognition and document transcription \cite{sam_writting, dulay2024informing, justin_tl}.
Much prior work has been done on image classification and biometrics, to the best of our knowledge, our contribution is the first to address the localization component of a detection model by introducing psychophysical measurements into the bounding box regression loss, while addressing the challenges of distance and occlusion from aerial views in ER scenarios.

%% file: sec_psycho_data.tex
\section{Psychophysical Dataset Collection}
\label{sec:psycho_data}

To collect behavioral measurements of how humans perform the task of person detection from occluded, aerial views, we created 500 different surveys of 13 questions, presenting a different aerial image on each question, with the instruction being ``Find the person in the picture".

\subsection{NOMAD's subsets selection}
NOMAD \cite{russell2024nomad} is a dataset of images and videos focused on ER scenarios, composed of 100 different actors filmed from a sUAS at five different distances (10m, 30m, 50m, 70m, and 90m).  Additionally, each bounding box of the dataset is annotated with a ``visibility" label (10 different levels), understood as the inverse of an occlusion metric.  We randomly selected a subset of NOMAD so that for every (distance, visibility) combination, we had one image of each actor.  The process gave us a subset of 4,883 images, as some actors did not provide images for all (distance, visibility) combinations.  Additionally, we randomly collected a second subset of 768 different images fairly easy to solve, to serve as control questions, with all images of this subset being of distance 10m and visibility either 100\% or 90\%. We will refer to the first subset as psych-NOMAD-positives, and to the second subset as psych-NOMAD-control.  Images from the subsets were assigned randomly to a survey so that every survey received three images to serve as control questions from the psych-NOMAD-control subset and 10 images from the psych-NOMAD-positives subset; these 10 images had the condition of being of different actor, different visibility label, and two images per distance label. All surveys contained 13 images in total.

\begin{figure}
  \centering 
  \begin{subfigure}{\linewidth}
    \includegraphics[width=\linewidth]{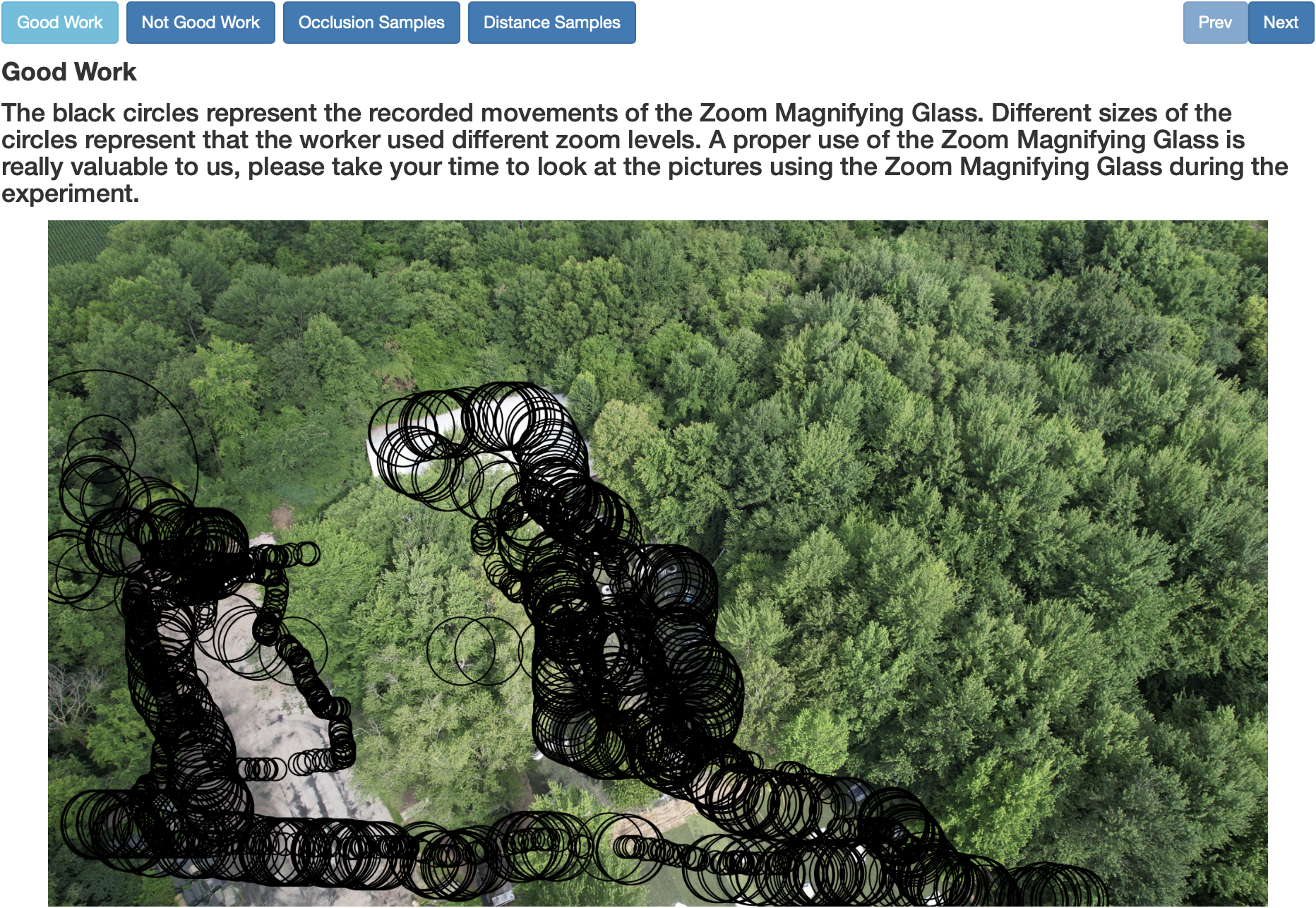}
    \caption{}
    \label{fig:good_sample}
  \end{subfigure} 
  \begin{subfigure}{\linewidth}
    \includegraphics[width=\linewidth]{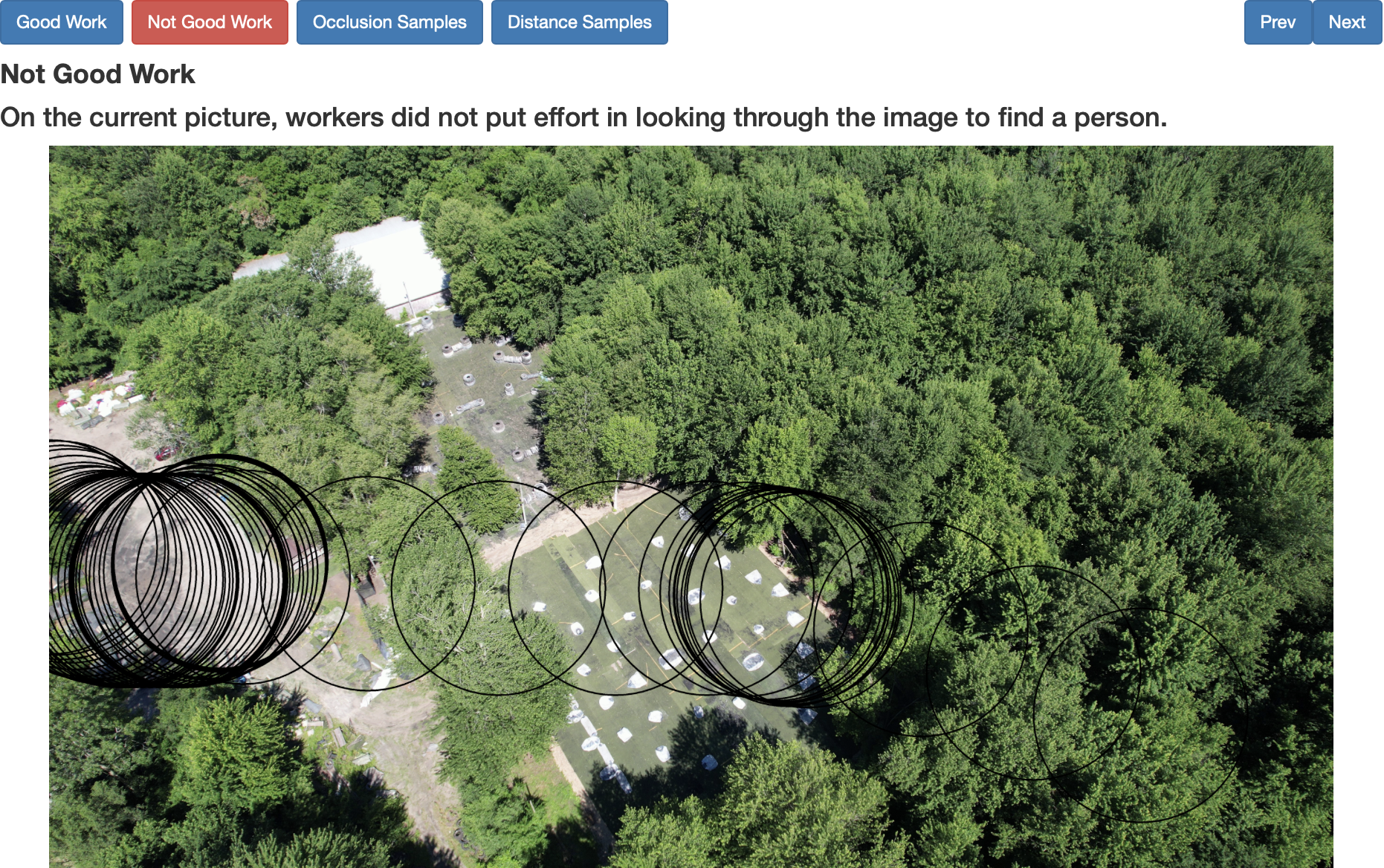}
    \caption{}
    \label{fig:bad_sample}
  \end{subfigure}
  \vspace{-0.5cm}
  \caption{Samples shown to workers prior to their survey experiment. The black circles represent the recorded location and area of the zoom-magnifying-glass. (a) Good work sample showing a thorough search. (b) Bad work sample, showing no search.}
  \vspace{-0.3cm}
  \label{fig:good_bad_samples}
\end{figure}

\begin{figure}[t]
  \centering
   \includegraphics[width=\linewidth]{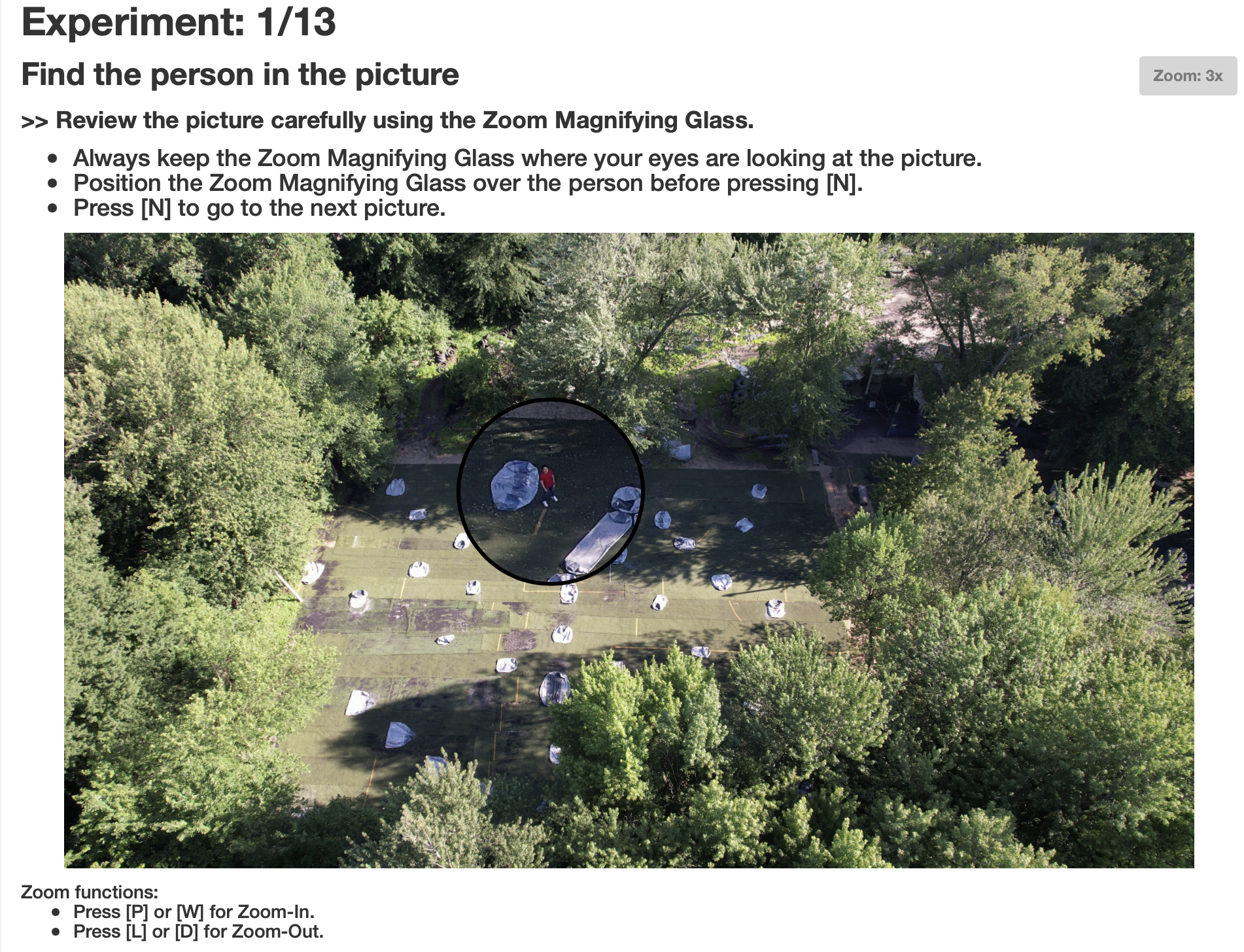}\vspace{-0.2cm}
   \caption{Display of the experiment setup, showing one image at a time, with reminders of key instructions.}
   \vspace{-0.5cm}
   \label{fig:experiment_sample}
\end{figure}

\begin{figure*}
  \centering
    \includegraphics[trim=0cm 0.5cm 0cm 2.5cm,clip,width=\linewidth]{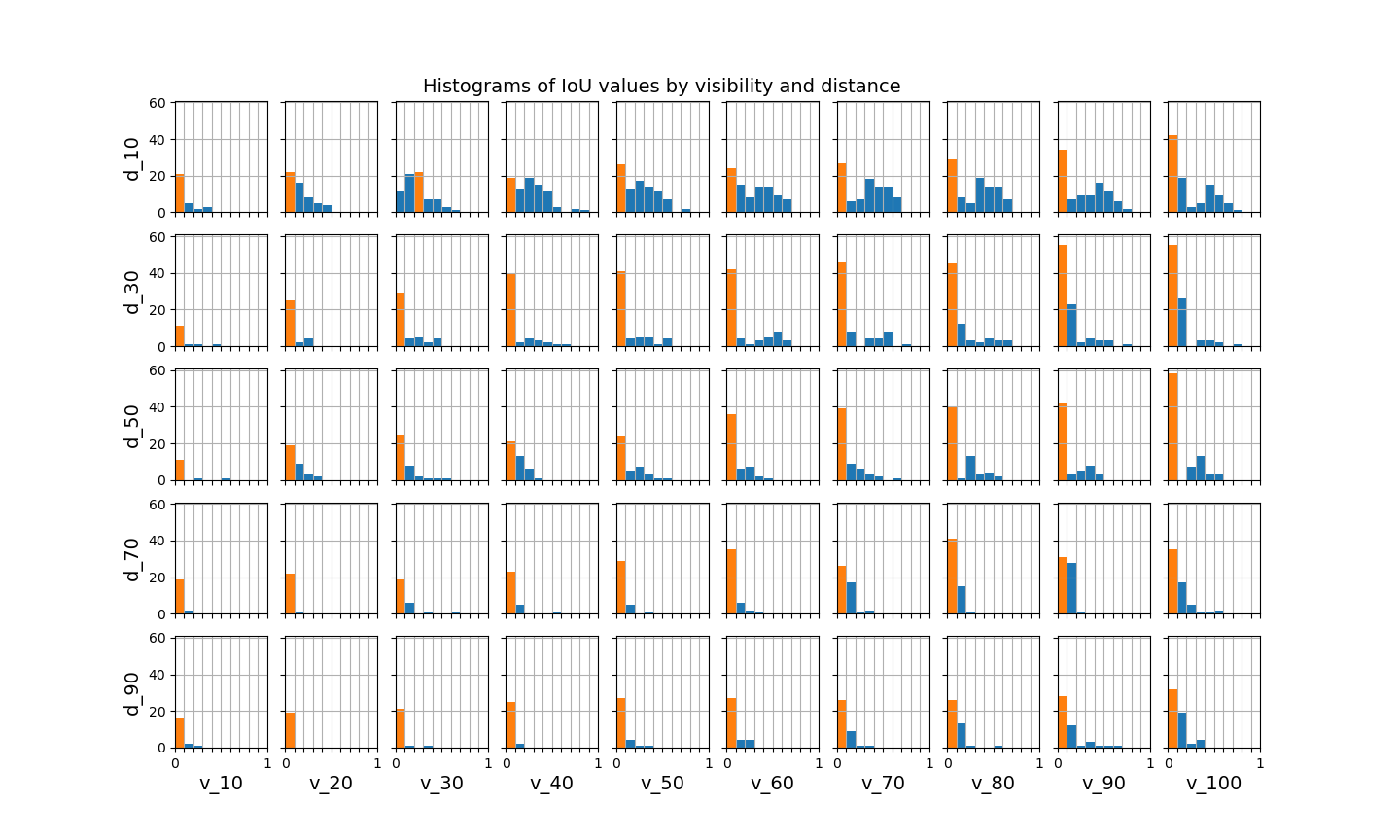}\vspace{-0.4cm}
  \caption{Histograms of IoU values $>$ 0, between the worker's area selection and the ground truth bounding box, with the smallest IoU-range bin of the histograms containing the most samples (highlighted bins), showing that workers were focusing on location rather than tightness of their area selection. The x-axis represents the visibility levels and the y-axis represents the distance levels.}\vspace{-0.5cm}
  \label{fig:iou_histogram}
\end{figure*}

\subsection{MTurk survey}
The MTurk surveys ``Find a person in the picture" were conducted under IRB protocol 18-01-4341.  We conducted a series of iterations and pre-tests to fine-tune the final survey process. The link provided in MTurk redirected the participants to our server application, where they were welcomed with an introduction, including short description, length of the survey, and approval and rejection criteria. The application included an online consent form that had to be acknowledged before proceeding to the survey instructions.

\textit{Survey instructions.} Participants were informed that every image in their survey contained a person.  They were instructed to carefully search through the image taking as much time as needed, requesting them to always keep their mouse cursor at the location of the image where they were looking at, and informing them that images were going to be evaluated by the quality of search (see \cref{fig:heatmap_sample}).  We provided them with a zoom-magnifying-glass attached to the cursor position over the image, as can be seen in \cref{fig:zoom}; once they were sure that their zoom-magnifying-glass was over the person in the image, they were to click the key [N] from their keyboard, taking them automatically to the next question.  After the instructions section, participants were shown a couple of samples of good and bad searches, as seen in \cref{fig:good_bad_samples}, as well as samples of increasing difficulty by occlusion and distance.  Finally, workers were presented with three practice images, simulating the actual experiment.  The practice images included easy and hard samples, by distance and occlusion. Only participants that successfully completed the practice were allowed to continue to the experiment; participants that failed the practice were able to retry it. The practice section was considered successful if and only if the participants selected an area with an intersection with the ground truth bounding box of the person in the image, for all three practice images.

\textit{Survey experiment.} \Cref{fig:experiment_sample} shows the setup for each image presented during the experiment.  Each worker solved 13 images, of which three were control images. Half-way through their survey they were provided with a score informing them of the number of correct answers so far, with the goal of motivating them to keep consistently searching for a person.  During the experiment, for every image we collected: (1) zoom-magnifying-glass information, including zoom level, position over the image, and time spent at every position; (2) accuracy, recorded using the last saved position of the zoom-magnifying-glass before the worker proceeded to the next picture; (3) response time, recording the time spent by a worker to solve the image.  Every survey submitted by a worker was reviewed through visualizations of their zoom-magnifying-glass data, response times and control questions accuracy.  Submissions that failed any of these criteria were rejected and their associated surveys were made available again for new workers.

\subsection{Data analysis}

\begin{figure}
  \centering 
  \begin{subfigure}{\linewidth}
    \includegraphics[trim=0cm 0.15cm 0cm 0.7cm,clip,width=\linewidth]{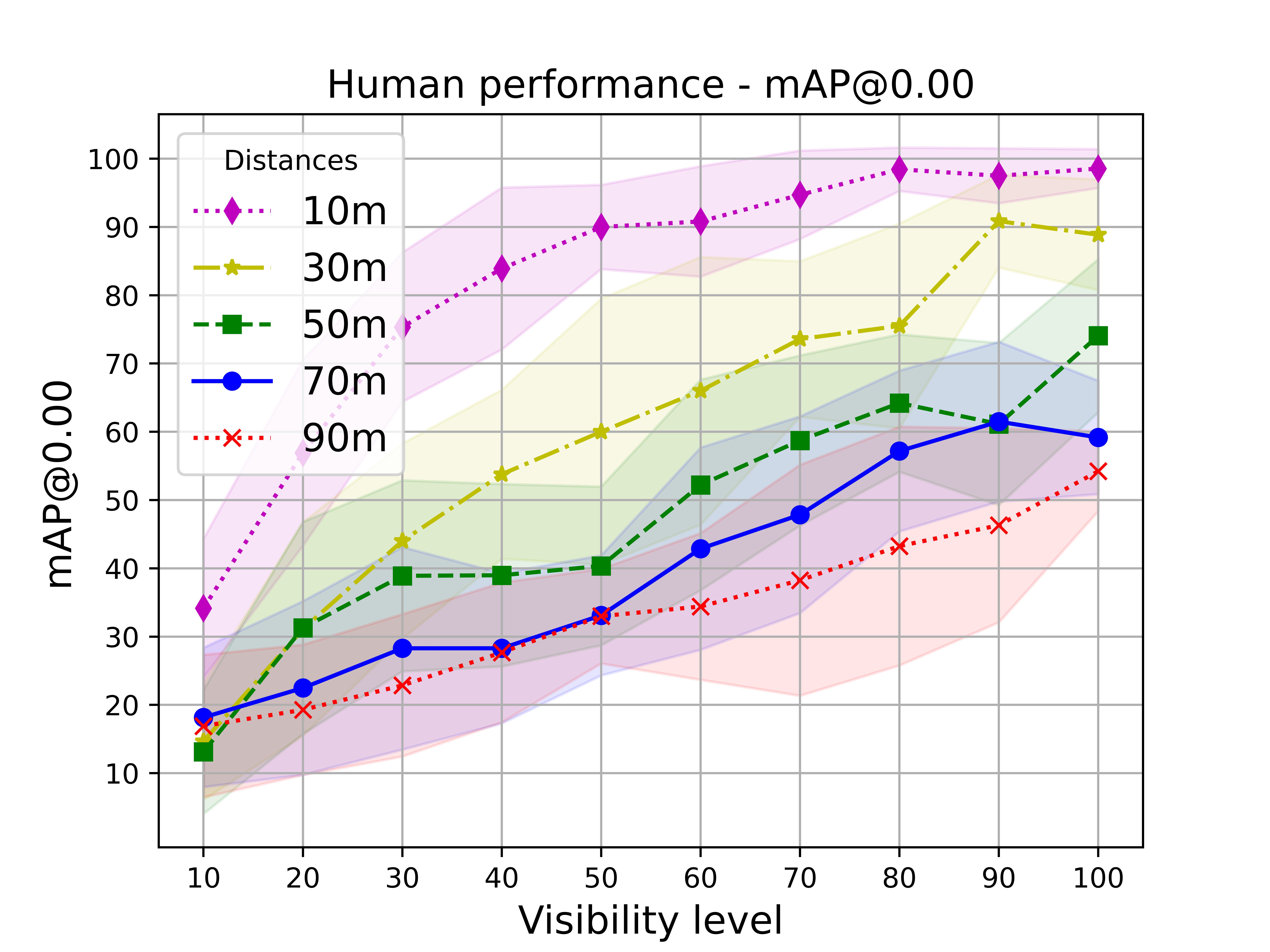}
    \caption{}
    \label{fig:human_performance_0}
  \end{subfigure} 
  \begin{subfigure}{\linewidth}
    \includegraphics[trim=0cm 0.15cm 0cm 0.7cm,clip,width=\linewidth]{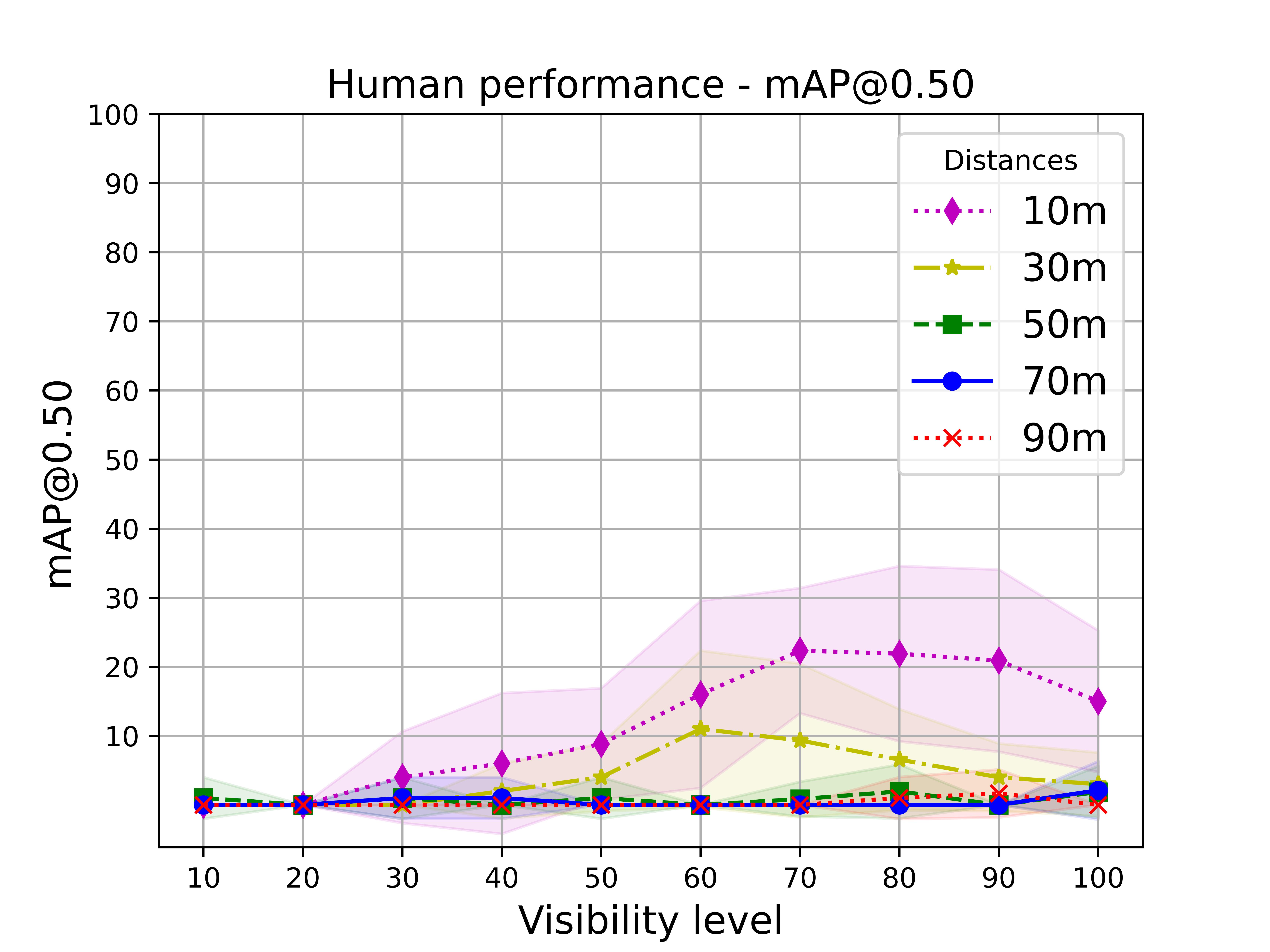}
    \caption{}
    \label{fig:human_performance_50}
  \end{subfigure}
  \vspace{-0.5cm}
  \caption{Human performance at (a) IoU $>$ 0 and (b) IoU $>$ 0.50, emphasizing that humans focus on location rather than the tightness of their area selection. Shaded areas show standard deviation from a 10 fold partition.}\vspace{0.2cm}
  \label{fig:human_performance}
\end{figure}

\begin{figure}[t]
  \centering
   \includegraphics[width=\linewidth]{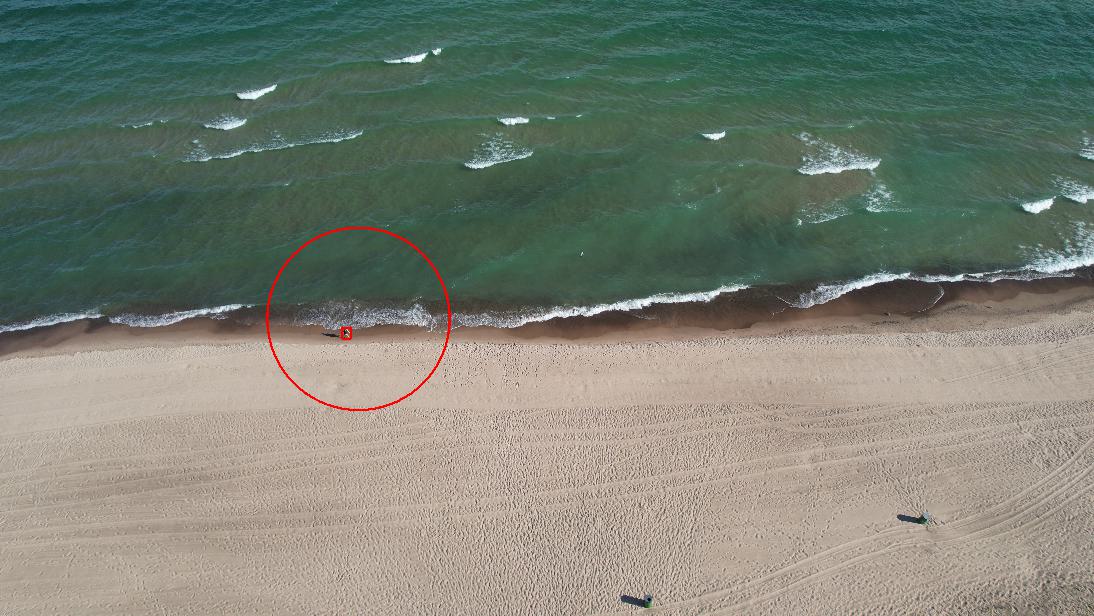}
   \caption{Sample picture from the experiment with an IoU of 0.0036867, exemplifying how humans did not care about tightness of their selection when given the instruction ``find the person in the picture". The red circle represents the worker's selection and the red square inside it represents the ground truth bounding box.}\vspace{-0.5cm}
   \label{fig:sample_low_IoU}
\end{figure}

\begin{figure}[t]
  \centering
   \includegraphics[trim=0cm 0cm 0cm 0.7cm,clip,width=\linewidth]{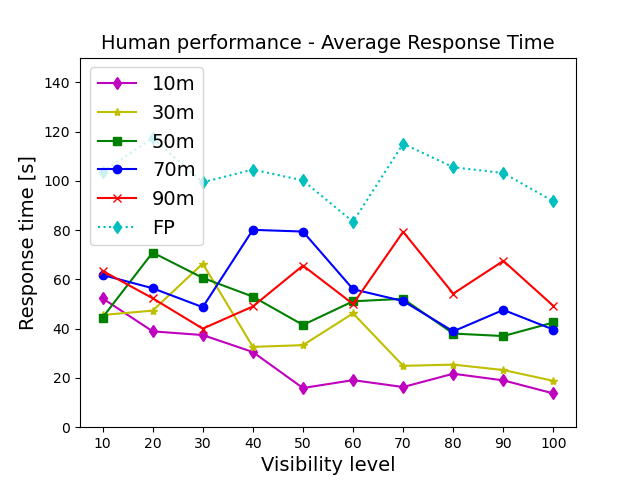}\vspace{-0.3cm}
   \caption{Average human response time for true positive samples across different distances and false positive samples, reporting a latency in the range of seconds for all plots, showing how CV models could benefit ER by performing inference in shorter times.}\vspace{0.2cm}
   \label{fig:human_latency}
\end{figure}

\begin{figure}[t]
  \centering
   \includegraphics[width=\linewidth]{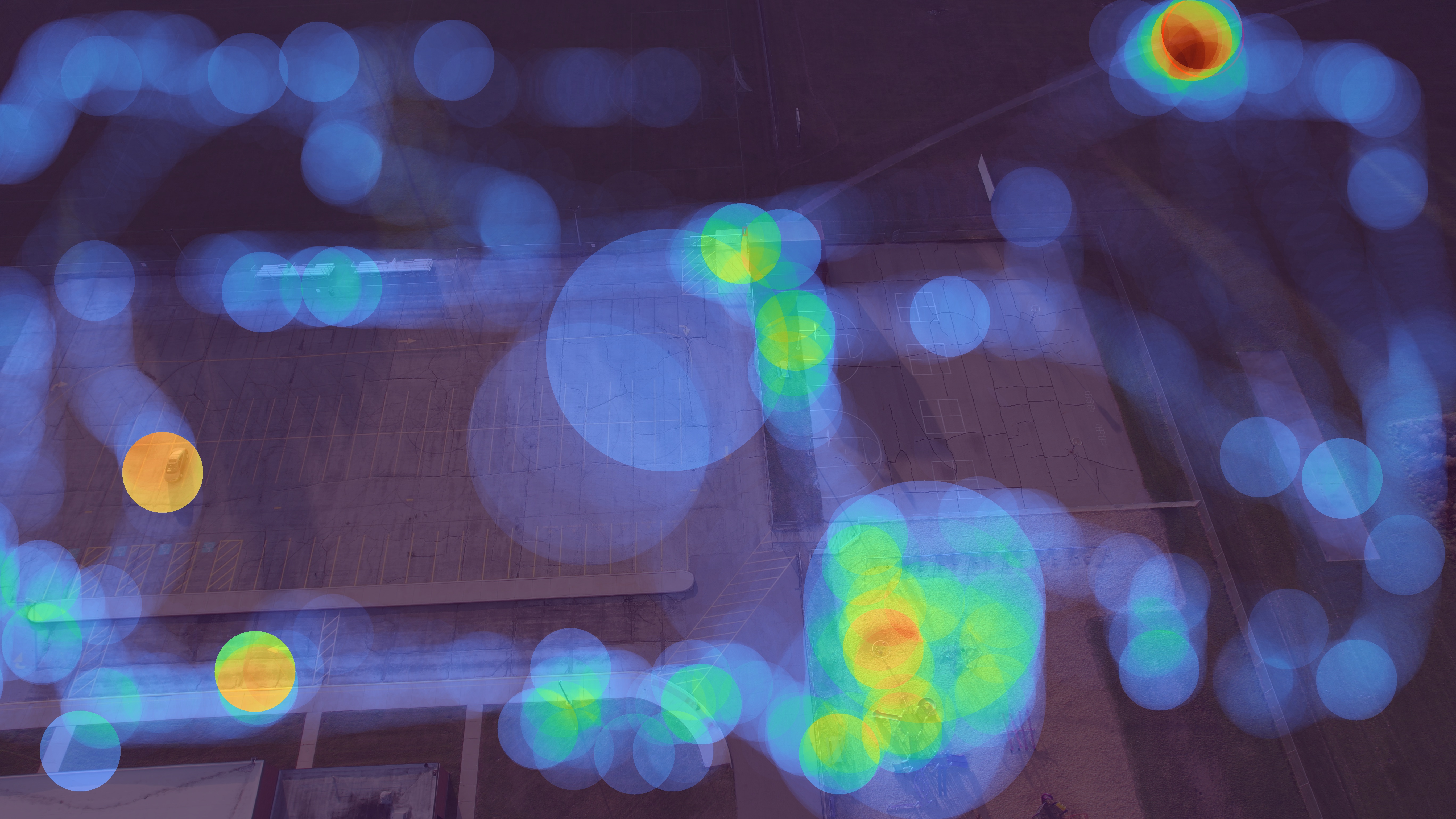}
   \caption{Sample search heatmap created from the location, area and time data of the worker's zoom-magnifying-glass.  The heatmap visualizations were used to ensure survey quality.}\vspace{-0.3cm}
   \label{fig:heatmap_sample}
\end{figure}

\textit{Accuracy.} We measured accuracy by taking the area of the last location over the image of the zoom-magnifying-glass circle, representing the worker's selection.  \Cref{fig:iou_histogram} shows a collection of histograms of 10 bins of the Intersection over Union (IoU) greater than zero between the worker's selection and the ground truth bounding box provided by NOMAD. From \cref{fig:iou_histogram}, we can distinguish two key observations: (1) human performance is hindered by increasing occlusion (decreasing visibility) and by increasing distance; (2) for most (distance, visibility) combinations, the highest frequency of samples is found at the smallest IoU-range bin. \Cref{fig:human_performance} shows human performance across different levels of distance and visibility, when evaluated with the mAP metric.  \Cref{fig:human_performance_0} shows results with the mAP measured with an IoU bigger than zero, while \cref{fig:human_performance_50} shows results with an IoU bigger than 0.50. From the above we observe that when humans are given the task of ``Find the person in the picture" without additional explicitness, they do not focus on the tightness of their selection, but rather on the person's location, as exemplified by \cref{fig:sample_low_IoU}.  

\textit{Response time.} \Cref{fig:human_latency} shows the average response time across different levels of distance and visibility for samples with IoU bigger than zero, considered as true positives, and the average response time for false positive samples. While we notice a significant time difference between false positives and true positives, both plots show that humans' response time is in the order of seconds, while CV models, such as YOLOv8 and RetinaNet run inference in the order of milliseconds \cite{yolov8_ultralytics, lin2017focal, detectron2}. ER scenarios are time critical situations, in an scenario where a human operator is searching for a missing person through the sUAS's camera and decides to fly at 30m, considering a ground sample distance of 12.729 mm/pixel \cite{russell2024nomad}, \cref{fig:human_latency} shows that it would take the human operator 18s to scan an area of 2500$m^2$, while CV detection models could take under 100ms \cite{detectron2modelzoo}, making this latency difference a key factor for the implementation and improvement of CV models onboard sUAS.

\textit{Search behaviour.} Additionally, the collected data from the zoom-magnifying-glass can be used to inform of human search behavior.  \Cref{fig:heatmap_sample} shows a sample visualization of the zoom-magnifying-glass data, using the area covered by different zoom levels and the time spent at each location to create a search heatmap.  During the data collection process, these visualizations were used to ensure search quality.

%% file: sec_psycho_loss.tex
\section{Psychometric Loss Formulation}
\label{sec:psycho_loss}

Our loss formulation is guided by the observation that without additional explicitness, humans interpret the instruction ``Find a person in the picture" as finding the (x,y) location of the person in the image, rather than finding a tight area to enclose the person in the image.  This observation is key and supports ER scenarios, where recall is critical.  Our first intuition is to guide the model with a penalty based on the deviation of the centers between prediction and ground truth.  Because this observation came from the psychophysical data, our second intuition is that the human behavioural data should guide how much this penalty should be.  We used inspiration from \cite{zhou2019objects} to build a normal probability density function around the center coordinates of the ground truth bounding box:
\begin{equation}
\label{eq:densityfunction}
\begin{split}
    &f = e^-\frac{(x_{pred} - x_{gt})^2 + (y_{pred} - y_{gt})^2}{2\sigma^2}
\end{split}
\end{equation}

Note that we are bounding the normal probability density function to be between 0 and 1.  As the variance factor determines how precise or disperse is our distribution, we use it to introduce the human behavioural data:

\vspace{-0.5cm}
\begin{equation}
\label{eq:variance}
\begin{split}
    &\sigma(d,v) = 100 - mAP@0.00(d,v)
\end{split}
\end{equation}
\vspace{-0.5cm}

\begin{figure}[t]
  \centering
   \includegraphics[width=\linewidth]{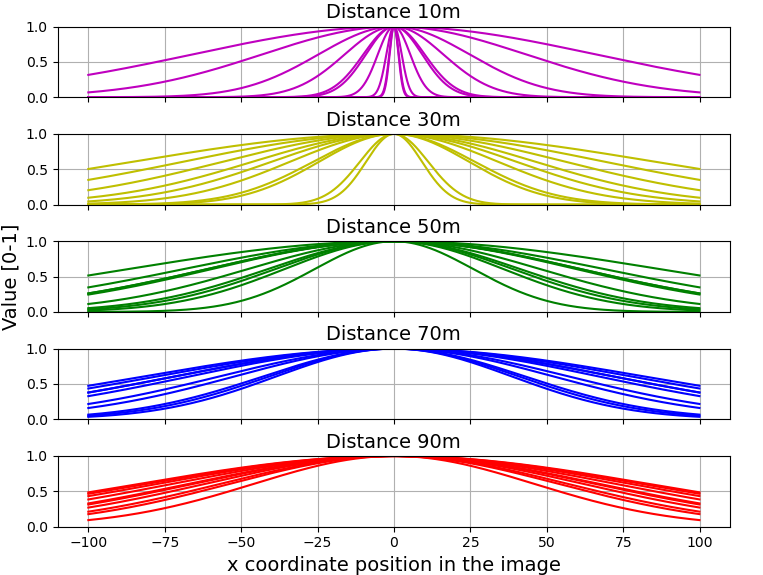}
   \caption{Normal probability density functions with the average human mAP@0.00 data as variance factor (see \cref{fig:human_performance_0}), to be introduced into the psychophysical loss function.  The inner functions represent higher levels of visibility. We can observe that for shorter distances and higher levels of visibility the penalty will be more strict, while it will allow smoother penalty progressions as we increase in distance and decrease in visibility. }\vspace{-0.4cm}
   \label{fig:probability_distribution}
\end{figure}

\Cref{fig:human_performance_0} shows the averaged values used as mAP@0.00 per distance and visibility. \Cref{fig:probability_distribution} shows the normal probability density functions with the human behavioural data as variance. From this we can observe that for shorter distances and higher levels of visibility the penalty will be more strict, while it will allow smoother penalty progressions as we increase in distance and occlusion.  As we want the penalty to decrease as the predicted center coordinates get closer to the ground truth center coordinates, the human penalty factor and the loss are expressed as:

\vspace{-0.4cm}
\begin{equation}
\label{eq:humanpenalty}
\begin{split}
    &human\_penalty(d,v) = \\
    & 1 - e^-\frac{(x_{pred} - x_{gt})^2 + (y_{pred} - y_{gt})^2}{2\sigma(d,v)^2}
\end{split}
\end{equation}
\vspace{-0.05cm}
\begin{equation}
\label{eq:humanloss}
\begin{split}
    &human\_loss(d,v) = A * human\_penalty(d,v) + \\
    & B * (1 - human\_penalty(d,v)) * default\_loss
\end{split}
\end{equation}

where A and B are weighting hyperparameters, and the $(1 - human\_penalty(d,v))$ factor forces the loss to concentrate initially on the (x,y) coordinates of the center of the bounding box, and it only starts to shift its focus towards tightening the bounding box prediction as the $human\_penalty$ decreases.  Finally, the $human\_penalty$ is only calculated for the training samples that were part of the MTurk experiment, otherwise it is set to 0.

%% file: sec_results.tex
\section{Results}
\label{sec:results}
\begin{figure*}
  \centering 
  \begin{subfigure}{.49\linewidth}
    \includegraphics[trim=0cm 0.15cm 0cm 0.8cm,clip,width=\linewidth]{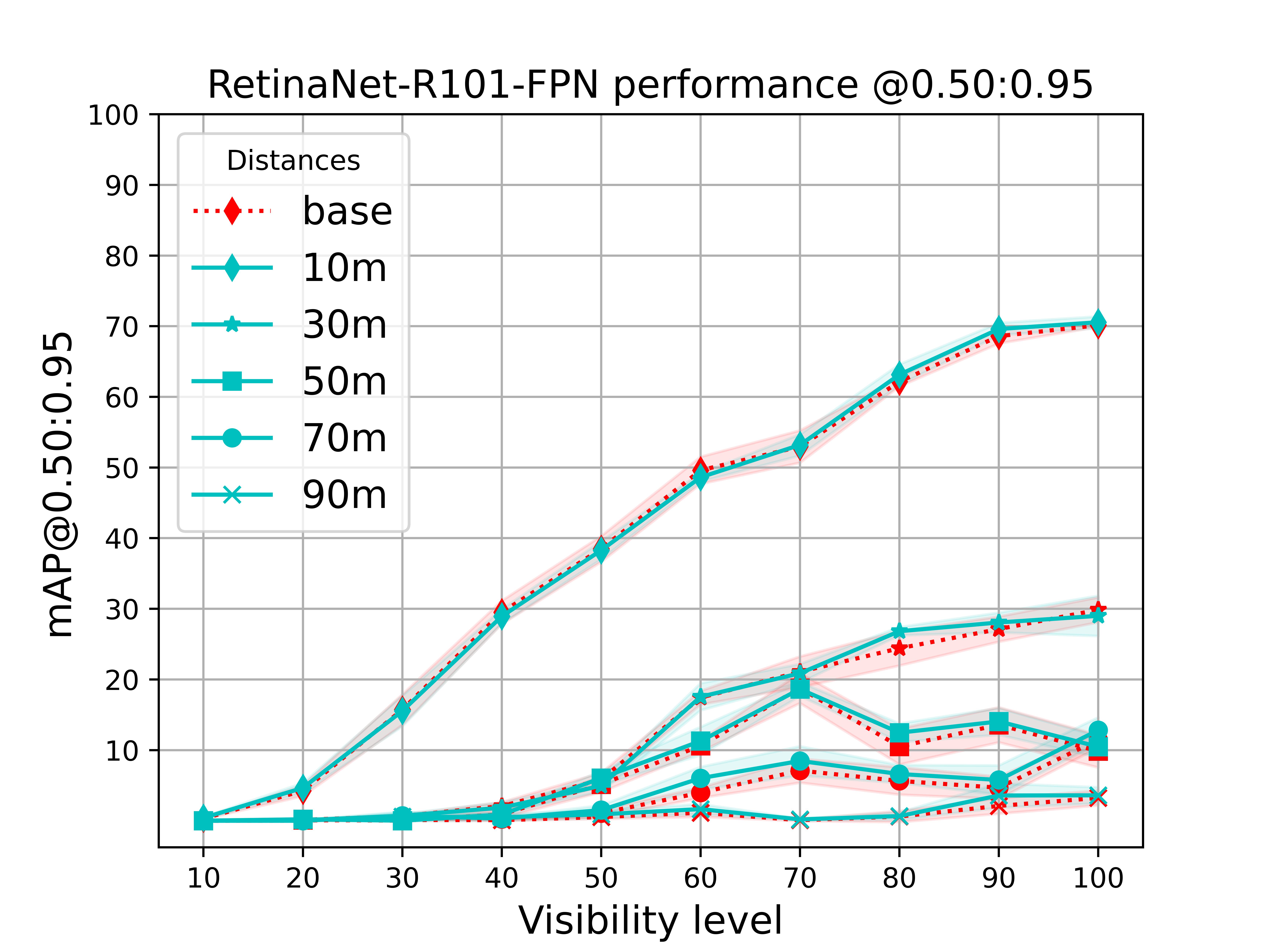}
    \caption{}
    \label{fig:retinanet_ap}
  \end{subfigure} 
  \begin{subfigure}{.49\linewidth}
    \includegraphics[trim=0cm 0.15cm 0cm 0.3cm,clip,width=\linewidth]{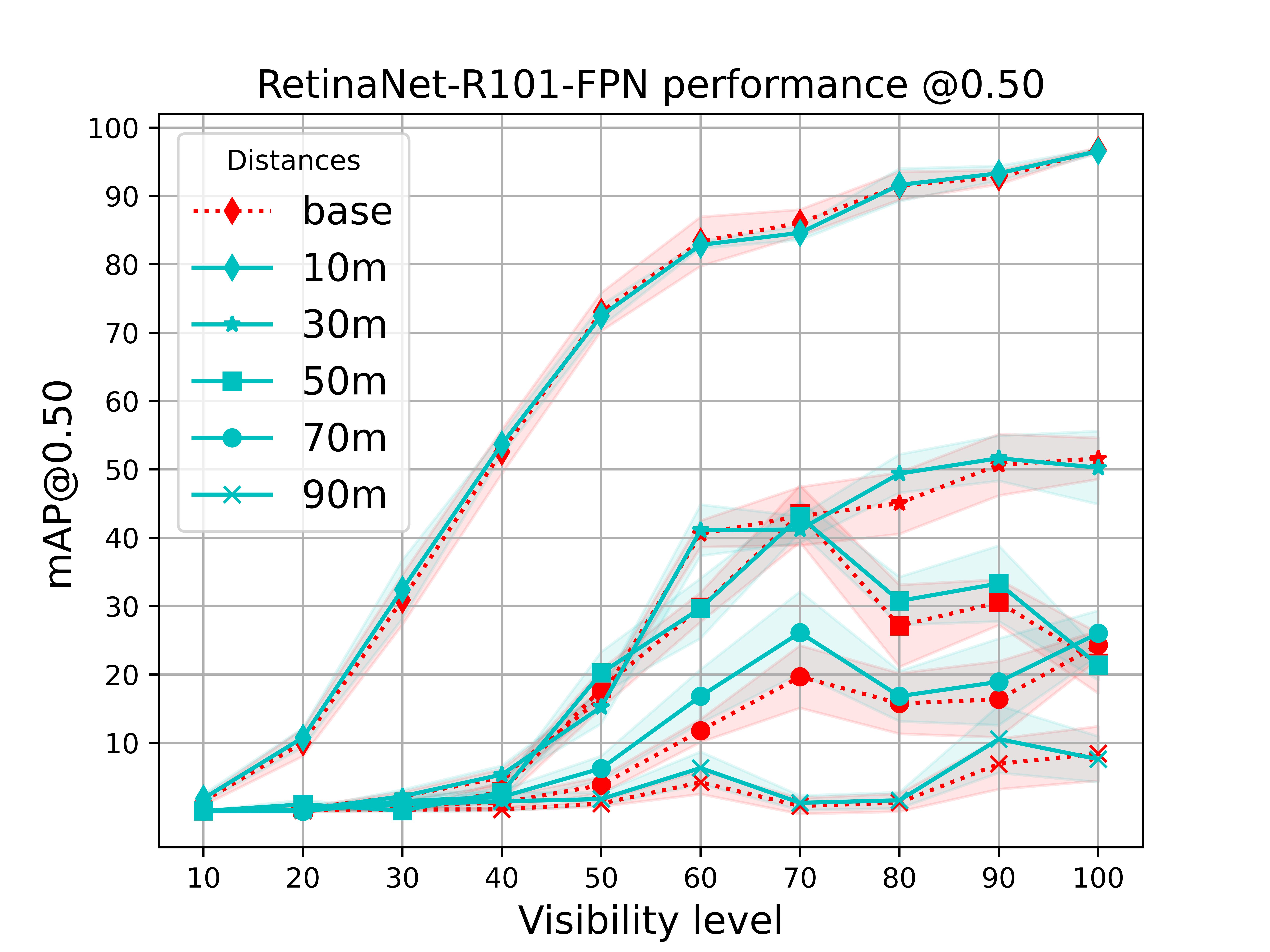}
    \caption{}
    \label{fig:retinanet_ap50}
  \end{subfigure}
  \vspace{-0.3cm}
  \caption{Performance of the RetinaNet-R101-FPN model against different levels of distance and occlusion, when trained and tested on NOMAD, and evaluated on (a) mAP@0.50:0.95 and (b) mAP@0.50.  The baseline model, shown in dashed lines, was trained with its default regression loss.  Our models using the psychophysical loss are shown in continuous lines.  Plot (b) shows how we improve the performance at higher distances (specifically at the distance of 70m) across different levels of occlusion, without hindering the performance at lower distances.  Additionally, plot (a) shows that although our psychophysical loss adaptation focuses on improving center location coordinates, which corresponds with looser IoU values reporting better improvements, it does not hinder the tightness of the bounding box. Our focus is on looser IoU evaluations, as location has been found to be more important than tightness of bounding box in ER scenarios.}\vspace{-0.4cm}
  \label{fig:retinanet}
\end{figure*}

To demonstrate the use of our formulated psychophysical loss, we used the RetinaNet-R101-FPN model from the Detectron2 library \cite{detectron2modelzoo}, and we created a random split of 80, 10, and 10 actors from the 100 actors of NOMAD, for training, validation and testing, respectively, representing a test set of roughly 4000 images, as NOMAD is composed of 42825 images in total. RetinaNet \cite{lin2017focal} is a CV model that has two heads, the first head performs bounding box regression using the SmoothL1 loss, while the second head performs classification using its focal loss. As our psychophysical loss adaptation addresses the location component of CV detection models, this represents the bounding box regression loss for the RetinaNet model.  To compare performance of our psychophysical loss, we trained the RetinaNet model on two different scenarios during 10 epochs: (1) the RetinaNet model with its default loss (baseline), (2) the RetinaNet model with our modified loss. Both scenarios were repeated five times to obtain statistical data, with all hyperparameters maintained the same for all runs; all iterations were done using three  Quadro RTX 6000 GPUs with 23040MiB of memory.  The weights A and B (see \cref{eq:humanloss}) of our psychophysical loss where set to 0.05 and 0.95 respectively, during all runs.

\Cref{fig:retinanet} shows the RetinaNet-R101-FPN model's performance, comparing the performance of our psychophysical loss against the baseline, across the different levels of distance and visibility (inverse of occlusion) of NOMAD.  \Cref{fig:retinanet_ap} evaluates the models with the mAP@0.50:0.95 metric, showing that although our psychophysical loss adaptation focuses on improving the center location coordinates of the bounding boxes, which corresponds with looser IoU values reporting better improvements, it does not hinder the tightness of the bounding box compared to the baseline. Our evaluation focuses in the mAP@0.50 rather than the mAP@0.50:0.95 metric, as the analysis of the human behavioral data showed that when humans are given the task of ``finding the person in the picture", without additional explicitness, they do not focus on the tightness of their selection, but rather on the person's location.  \Cref{fig:retinanet_ap50} evaluates the models with the mAP@0.50 metric, showing improvements over the baseline at higher distances, specifically at the distance of 70m, and across different levels of occlusion, without hindering the performance at lower distances. As person recognition at lower distances (\eg 10m) is relatively easier, we would not expect much improvement compared to the baseline, nonetheless, as the level of difficulty increases with further distances, there is more room for improvement, and greater benefits are observed.  However, our loss adaptation exhibits a peak improvement against the baseline at 70m, with the difficulty levels at 90m representing a bottleneck for our loss function as well. It is worth noting that our approach incurs minimal additional computational cost during training and does not incur any additional computational cost during inference, this being important as CV models are intended to be deployed on sUAS with limited computational resources.

%% file: sec_futurework_conclusions.tex
\section{Future Work and Conclusions}
\label{sec:future_work_conclusions}

The work in this paper has presented the usefulness of Psych-ER, a human behavioral dataset collected for person detection from aerial, occluded views, for Emergency Response (ER) scenarios. Future work will continue to exploit this behavioral dataset with the goal of improving Computer Vision (CV) performance, focusing on: 

\textit{Human behavioral dataset}: so far we have only explored the use of the human accuracy data on RetinaNet.  Future work aims to exploit all the behavioral data collected, including response time and search behaviour, testing several CV architectures.

\textit{Custom CV architecture}: besides adapting existing architectures, our goal is to create a custom CV architecture for ER scenarios, incorporating psychophysical measurements and full context of the application, such as embedding information available through the sUAS (\eg flight altitude).

\textit{Onboard CV pipeline}: sUAS systems could have access to several CV models trained to tackle aerial detection at specific altitudes, given evidence of the effectiveness of our loss function, future work is to explore training CV models with our psychophysical loss for specific altitudes.

\textit{Real-world deployment}: we have found through our own experiences in deploying CV on sUAS that there are always unforeseen additional challenges when moving to the real-world. Our final objective is to deploy and evaluate ER-tailored CV models on physical sUAS.

From \cref{fig:retinanet} we can observe that occlusion remains a key factor that hinders performance of CV models, as performance is too poor for visibility scores smaller than 50\% and distances greater than 10m.  In the realm of real-world deployment of sUAS during ER missions, the victim’s position is often within the field of view for several seconds during a fly-over, opening the door to explore the benefits of analyzing contiguous image frames to create opportunities to detect the victim from increasing levels of visibility, additionally, sUAS software systems can be created to react according to different levels of confidence of the detection, where a low-confidence sighting could cause  the sUAS to adapt its flight path to decrease altitude and attempt to optimize the view so as to increase visibility and either confirm or rule out the sighting. The integration of sUAS with onboard CV capabilities into ER scenarios represents a key component for the improvement of the overall efficiency of ER missions, where mission success may represent life and death. The contributions of this paper are: (1) the human behavioral dataset Psych-ER, containing human search behaviour of more than 5000 aerial images from the NOMAD dataset, an ER-focused dataset, and (2) a psychophysical loss formulation addressing the location component of CV detection models using the accuracy data from Psych-ER.  Our future steps will continue to exploit the human behavioral data with the endmost goal of deploying and evaluating ER-tailored CV models on physical sUAS.

%% file: main.bbl
\begin{thebibliography}{10}\itemsep=-1pt

\bibitem{sophia_reliability_score}
Sophia~J. Abraham, Zachariah Carmichael, Sreya Banerjee, Rosaura~G. VidalMata, Ankit Agrawal, Md~Nafee~Al Islam, Walter~J. Scheirer, and Jane Cleland{-}Huang.
\newblock Adaptive autonomy in human-on-the-loop vision-based robotics systems.
\newblock In {\em 1st {IEEE/ACM} Workshop on {AI} Engineering - Software Engineering for AI, WAIN@ICSE 2021, Madrid, Spain, May 30-31, 2021}, pages 113--120. {IEEE}, 2021.

\bibitem{chi20-partnerships}
Ankit Agrawal, Sophia~J. Abraham, Benjamin Burger, Chichi Christine, Luke Fraser, John~M. Hoeksema, Sarah Hwang, Elizabeth Travnik, Shreya Kumar, Walter~J. Scheirer, Jane Cleland{-}Huang, Michael Vierhauser, Ryan Bauer, and Steve Cox.
\newblock The next generation of human-drone partnerships: Co-designing an emergency response system.
\newblock In Regina Bernhaupt, Florian~'Floyd' Mueller, David Verweij, Josh Andres, Joanna McGrenere, Andy Cockburn, Ignacio Avellino, Alix Goguey, Pernille Bj{\o}n, Shengdong Zhao, Briane~Paul Samson, and Rafal Kocielnik, editors, {\em {CHI} '20: {CHI} Conference on Human Factors in Computing Systems, Honolulu, HI, USA, April 25-30, 2020}, pages 1--13. {ACM}, 2020.

\bibitem{droneresponse}
Md~Nafee Al~Islam, Muhammed~Tawfiq Chowdhury, Ankit Agrawal, Michael Murphy, Raj Mehta, Daria Kudriavtseva, Jane Cleland-Huang, Michael Vierhauser, and Marsha Chechik.
\newblock Configuring mission-specific behavior in a product line of collaborating small unmanned aerial systems.
\newblock {\em Journal of Systems and Software}, 197:111543, 2023.

\bibitem{aerial_trackingSignal}
Antonio Albanese, Vincenzo Sciancalepore, and Xavier Costa-Pérez.
\newblock hierarquicallyo: An automated search-and-rescue drone-based solution for victims localization.
\newblock {\em IEEE Transactions on Mobile Computing}, 21(9):3312--3325, 2022.

\bibitem{lowpowercv}
Sergei Alyamkin, Matthew Ardi, Alexander~C Berg, Achille Brighton, Bo Chen, Yiran Chen, Hsin-Pai Cheng, Zichen Fan, Chen Feng, Bo Fu, et~al.
\newblock Low-power computer vision: Status, challenges, and opportunities.
\newblock {\em IEEE Journal on Emerging and Selected Topics in Circuits and Systems}, 9(2):411--421, 2019.

\bibitem{multires_drones}
Giuseppe Amato, Fabrizio Falchi, Claudio Gennaro, Fabio~Valerio Massoli, and Claudio Vairo.
\newblock Multi-resolution face recognition with drones.
\newblock In Vit Vozenilek, editor, {\em {SSIP} 2020: 2020 3rd International Conference on Sensors, Signal and Image Processing, Prague, Czech Republic, October 9-11, 2020}, pages 13--18. {ACM}, 2020.

\bibitem{pedestrian_multimodal}
Kirsnaragavan Arudpiragasam, Taraka Rama Krishna~Kanth Kannuri, Klaus Schwarz, Michael Hartmann, and Reiner Creutzburg.
\newblock Real-time pedestrian detection using radar-camera fusion and clustering.
\newblock In {\em Multimodal Image Exploitation and Learning 2023}, volume 12526, pages 89--103. SPIE, 2023.

\bibitem{bendale2015towards}
Abhijit Bendale and Terrance Boult.
\newblock Towards open world recognition.
\newblock In {\em Proceedings of the IEEE conference on computer vision and pattern recognition}, pages 1893--1902, 2015.

\bibitem{airsight}
Arturo Miguel~Russell Bernal and Jane Cleland-Huang.
\newblock Hierarchically organized computer vision in support of multi-faceted search for missing persons.
\newblock In {\em 2023 IEEE 17th International Conference on Automatic Face and Gesture Recognition (FG)}, pages 1--7. IEEE, 2023.

\bibitem{yolov4}
Alexey Bochkovskiy, Chien-Yao Wang, and Hong-Yuan~Mark Liao.
\newblock Yolov4: Optimal speed and accuracy of object detection.
\newblock {\em arXiv preprint arXiv:2004.10934}, 2020.

\bibitem{boyd2022human}
Aidan Boyd, Kevin~W Bowyer, and Adam Czajka.
\newblock Human-aided saliency maps improve generalization of deep learning.
\newblock In {\em Proceedings of the IEEE/CVF Winter Conference on Applications of Computer Vision}, pages 2735--2744, 2022.

\bibitem{adam_new}
Aidan Boyd, Daniel Moreira, Andrey Kuehlkamp, Kevin Bowyer, and Adam Czajka.
\newblock Human saliency-driven patch-based matching for interpretable post-mortem iris recognition.
\newblock In {\em Proceedings of the IEEE/CVF Winter Conference on Applications of Computer Vision (WACV) Workshops}, pages 701--710, January 2023.

\bibitem{boyd2023cyborg}
Aidan Boyd, Patrick Tinsley, Kevin~W Bowyer, and Adam Czajka.
\newblock Cyborg: Blending human saliency into the loss improves deep learning-based synthetic face detection.
\newblock In {\em Proceedings of the IEEE/CVF Winter Conference on Applications of Computer Vision}, pages 6108--6117, 2023.

\bibitem{custom_rcnn_heridal}
Dunja Bozic{-}Stulic, Zeljko Marusic, and Sven Gotovac.
\newblock Deep learning approach in aerial imagery for supporting land search and rescue missions.
\newblock {\em Int. J. Comput. Vis.}, 127(9):1256--1278, 2019.

\bibitem{wisard}
Daniel Broyles, Christopher~R. Hayner, and Karen Leung.
\newblock Wisard: A labeled visual and thermal image dataset for wilderness search and rescue.
\newblock In {\em 2022 IEEE/RSJ International Conference on Intelligent Robots and Systems (IROS)}, pages 9467--9474, 2022.

\bibitem{news_teamwork}
{Brunswick County Sheriff's Office, NC}.
\newblock Teamwork makes the dream work, 2022.
\newblock \url{https://www.facebook.com/brunswicksheriffNC/posts/355462306611443}, Last accessed on 2024-09-09.

\bibitem{hci_murphy}
J. Casper and R.R. Murphy.
\newblock Human-robot interactions during the robot-assisted urban search and rescue response at the world trade center.
\newblock {\em IEEE Transactions on Systems, Man, and Cybernetics, Part B (Cybernetics)}, 33(3):367--385, 2003.

\bibitem{news_fire}
{CBS NEWS}.
\newblock Italy police use drone to catch suspected arsonist in the act as wildfires char calabria, 2023.
\newblock \url{https://www.cbsnews.com/news/italy-fire-drone-arson-suspect-arrest-calabria-wildfires-europe-algeria/}, Last accessed on 2024-09-09.

\bibitem{chambers2024self}
Theodore Chambers, Jane Cleland-Huang, and Michael Vierhauser.
\newblock Self-adaptation of loosely coupled systems across a system of small uncrewed aerial systems.
\newblock In {\em Proceedings of the 12th ACM/IEEE International Workshop on Software Engineering for Systems-of-Systems and Software Ecosystems}, pages 37--44, 2024.

\bibitem{chambers2024hifuzz}
Theodore Chambers, Michael Vierhauser, Ankit Agrawal, Michael Murphy, Jason~Matthew Brauer, Salil Purandare, Myra~B Cohen, and Jane Cleland-Huang.
\newblock Hifuzz: Human interaction fuzzing for small unmanned aerial vehicles.
\newblock In {\em Proceedings of the CHI Conference on Human Factors in Computing Systems}, pages 1--14, 2024.

\bibitem{occlusion_oldsurvey}
Himanshu Chandel and Sonia Vatta.
\newblock Occlusion detection and handling: a review.
\newblock {\em International Journal of Computer Applications}, 120(10), 2015.

\bibitem{aerial_ground_collaboration}
Dimitrios Chatziparaschis, Michail~G Lagoudakis, and Panagiotis Partsinevelos.
\newblock Aerial and ground robot collaboration for autonomous mapping in search and rescue missions.
\newblock {\em Drones}, 4(4):79, 2020.

\bibitem{survey_low_resolution}
Guang Chen, Haitao Wang, Kai Chen, Zhijun Li, Zida Song, Yinlong Liu, Wenkai Chen, and Alois Knoll.
\newblock A survey of the four pillars for small object detection: Multiscale representation, contextual information, super-resolution, and region proposal.
\newblock {\em IEEE Transactions on systems, man, and cybernetics: systems}, 52(2):936--953, 2020.

\bibitem{occlusion_humanperception}
Xiaowu Chen, Qing Li, Dongyue Zhao, and Qinping Zhao.
\newblock Occlusion cues for image scene layering.
\newblock {\em Computer Vision and Image Understanding}, 117(1):42--55, 2013.

\bibitem{challenges_tawfiq}
Muhammed~Tawfiq Chowdhury and Jane Cleland-Huang.
\newblock Engineering challenges for ai-supported computer vision in small uncrewed aerial systems.
\newblock In {\em 2023 IEEE/ACM 2nd International Conference on AI Engineering – Software Engineering for AI (CAIN)}, pages 158--170, 2023.

\bibitem{hmt_Cleland-HuangAV22}
Jane Cleland{-}Huang, Ankit Agrawal, Michael Vierhauser, Michael Murphy, and Mike Prieto.
\newblock Extending {MAPE-K} to support human-machine teaming.
\newblock In Bradley~R. Schmerl, Martina Maggio, and Javier C{\'{a}}mara, editors, {\em International Symposium on Software Engineering for Adaptive and Self-Managing Systems, {SEAMS} 2022, Pittsburgh, PA, USA, May 22-24, 2022}, pages 120--131. {ACM/IEEE}, 2022.

\bibitem{hmt_cleland2024human}
Jane Cleland-Huang, Theodore Chambers, Sebastian Zudaire, Muhammed~Tawfiq Chowdhury, Ankit Agrawal, and Michael Vierhauser.
\newblock Human--machine teaming with small unmanned aerial systems in a mape-k environment.
\newblock {\em ACM Transactions on Autonomous and Adaptive Systems}, 19(1):1--35, 2024.

\bibitem{news_drowning}
{CNN}.
\newblock Drone lifeguard saves 14-year-old from drowning, 2022.
\newblock \url{https://www.cnn.com/videos/us/2022/07/25/drone-rescues-teenager-drowning-spain-orig-cb-aw.cnn}, Last accessed on 2024-09-09.

\bibitem{crum2023teaching}
Colton~R Crum, Aidan Boyd, Kevin Bowyer, and Adam Czajka.
\newblock Teaching ai to teach: Leveraging limited human salience data into unlimited saliency-based training.
\newblock {\em arXiv preprint arXiv:2306.05527}, 2023.

\bibitem{crum2023mentor}
Colton~R Crum and Adam Czajka.
\newblock Mentor: Human perception-guided pretraining for iris presentation detection.
\newblock {\em arXiv preprint arXiv:2310.19545}, 2023.

\bibitem{adam_iris}
Adam Czajka, Daniel Moreira, Kevin Bowyer, and Patrick Flynn.
\newblock Domain-specific human-inspired binarized statistical image features for iris recognition.
\newblock In {\em 2019 IEEE Winter Conference on Applications of Computer Vision (WACV)}, pages 959--967, 2019.

\bibitem{news_hikingAfrica}
{Drones.R.Africa}.
\newblock Drone helps save hiker on table mountain, cape town, 2019.
\newblock \url{https://dronenews.africa/drone-helps-save-man-on-table-mountain-cape-town/}, Last accessed on 2024-09-09.

\bibitem{duan2019centernet}
Kaiwen Duan, Song Bai, Lingxi Xie, Honggang Qi, Qingming Huang, and Qi Tian.
\newblock Centernet: Keypoint triplets for object detection.
\newblock In {\em Proceedings of the IEEE/CVF international conference on computer vision}, pages 6569--6578, 2019.

\bibitem{dulay2024informing}
Justin Dulay, Sonia Poltoratski, Till~S Hartmann, Samuel~E Anthony, and Walter~J Scheirer.
\newblock Informing machine perception with psychophysics.
\newblock {\em Proceedings of the IEEE}, 112(2):88--96, 2024.

\bibitem{justin_tl}
Justin Dulay and Walter~J Scheirer.
\newblock Using human perception to regularize transfer learning.
\newblock {\em arXiv preprint arXiv:2211.07885}, 2022.

\bibitem{news_earthquake}
{EL PAÍS}.
\newblock Gallery aerial images reveal scale of earthquake devastation in turkey and syria, 2023.
\newblock \url{https://english.elpais.com/international/2023-02-09/gallery-aerial-images-reveal-scale-of-earthquake-devastation-in-turkey-and-syria.html}, Last accessed on 2024-09-09.

\bibitem{news_lostWoods}
{EMS1}.
\newblock Drone helps {N.C.} first responders rescue missing teen, 2023.
\newblock \url{https://www.ems1.com/drones/articles/drone-helps-nc-first-responders-rescue-missing-teen-igX40AfVSNKreUJk/}, Last accessed on 2024-09-09.

\bibitem{ground_gasNose}
Han Fan, Victor Hernandez~Bennetts, Erik Schaffernicht, and Achim~J Lilienthal.
\newblock Towards gas discrimination and mapping in emergency response scenarios using a mobile robot with an electronic nose.
\newblock {\em Sensors}, 19(3):685, 2019.

\bibitem{walter_brain}
Ruth~C Fong, Walter~J Scheirer, and David~D Cox.
\newblock Using human brain activity to guide machine learning.
\newblock {\em Scientific reports}, 8(1):5397, 2018.

\bibitem{edge_yolo_wacv}
Prakhar Ganesh, Yao Chen, Yin Yang, Deming Chen, and Marianne Winslett.
\newblock Yolo-ret: Towards high accuracy real-time object detection on edge gpus.
\newblock In {\em Proceedings of the IEEE/CVF Winter Conference on Applications of Computer Vision (WACV)}, pages 3267--3277, January 2022.

\bibitem{gescheider2013psychophysics}
George~A Gescheider.
\newblock {\em Psychophysics: the fundamentals}.
\newblock Psychology Press, 2013.

\bibitem{news_oldadult}
{Global News}.
\newblock Police drone locates missing 81-year-old woman in north carolina, 2017.
\newblock \url{https://globalnews.ca/news/3851197/police-drone-locates-missing-81-year-old-woman-in-north-carolina/}, Last accessed on 2024-09-09.

\bibitem{granadeno2024environmentally}
Pedro Antonio~Alarcon Granadeno, Arturo Miguel~Russell Bernal, Md~Nafee Al~Islam, and Jane Cleland-Huang.
\newblock An environmentally complex requirement for safe separation distance between uavs.
\newblock In {\em 2024 IEEE 32nd International Requirements Engineering Conference Workshops (REW)}, pages 166--175. IEEE, 2024.

\bibitem{sam_writting}
Samuel Grieggs, Bingyu Shen, Greta Rauch, Pei Li, Jiaqi Ma, David Chiang, Brian Price, and Walter~J. Scheirer.
\newblock Measuring human perception to improve handwritten document transcription.
\newblock {\em IEEE Transactions on Pattern Analysis and Machine Intelligence}, 44(10):6594--6601, 2022.

\bibitem{huang2023measuring}
Jin Huang, Derek Prijatelj, Justin Dulay, and Walter Scheirer.
\newblock Measuring human perception to improve open set recognition.
\newblock {\em IEEE Transactions on Pattern Analysis and Machine Intelligence}, 45(9):11382--11389, 2023.

\bibitem{islam2024adam}
Md~Nafee~Al Islam, Jane Cleland-Huang, and Michael Vierhauser.
\newblock Adam: Adaptive monitoring of runtime anomalies in small uncrewed aerial systems.
\newblock In {\em Proceedings of the 19th International Symposium on Software Engineering for Adaptive and Self-Managing Systems}, pages 44--55, 2024.

\bibitem{yolov8_ultralytics}
Glenn Jocher, Ayush Chaurasia, and Jing Qiu.
\newblock Ultralytics {YOLOv8}, 2023.
\newblock \url{https://github.com/ultralytics/ultralytics}, Last accessed on 2024-09-09.

\bibitem{OWOD_CVPR_birth}
K~J Joseph, Salman Khan, Fahad~Shahbaz Khan, and Vineeth~N Balasubramanian.
\newblock Towards open world object detection.
\newblock In {\em Proceedings of the IEEE/CVF Conference on Computer Vision and Pattern Recognition (CVPR)}, pages 5830--5840, June 2021.

\bibitem{ground_snake}
Tetsushi Kamegawa, Taichi Akiyama, Satoshi Sakai, Kento Fujii, Kazushi Une, Eitou Ou, Yuto Matsumura, Toru Kishutani, Eiji Nose, Yusuke Yoshizaki, et~al.
\newblock Development of a separable search-and-rescue robot composed of a mobile robot and a snake robot.
\newblock {\em Advanced Robotics}, 34(2):132--139, 2020.

\bibitem{news_girl}
{KPRC 2 Click2Houston}.
\newblock Deputies use drone to find 3-year-old missing in washington brush, 2023.
\newblock \url{https://www.youtube.com/watch?v=tlwG-f_lRnM}, Last accessed on 2024-09-09.

\bibitem{law2018cornernet}
Hei Law and Jia Deng.
\newblock Cornernet: Detecting objects as paired keypoints.
\newblock In {\em Proceedings of the European conference on computer vision (ECCV)}, pages 734--750, 2018.

\bibitem{object_afc_edge_yolo}
Jeonghun Lee and Kwang-il Hwang.
\newblock Yolo with adaptive frame control for real-time object detection applications.
\newblock {\em Multimedia Tools and Applications}, 81(25):36375--36396, 2022.

\bibitem{news_firehomes}
{Lehighvalleylive}.
\newblock {F}erry {S}treet fire update: 10 homes damaged need to be demolished, cause still probed, 2023.
\newblock \url{https://www.lehighvalleylive.com/easton/2023/07/ferry-street-fire-update-10-homes-damaged-need-to-be-demolished-cause-still-probed.html}, Last accessed on 2024-09-09.

\bibitem{lin2017focal}
Tsung-Yi Lin, Priya Goyal, Ross Girshick, Kaiming He, and Piotr Doll{\'a}r.
\newblock Focal loss for dense object detection.
\newblock In {\em Proceedings of the IEEE international conference on computer vision}, pages 2980--2988, 2017.

\bibitem{linsley2018learning}
Drew Linsley, Dan Shiebler, Sven Eberhardt, and Thomas Serre.
\newblock Learning what and where to attend.
\newblock {\em arXiv preprint arXiv:1805.08819}, 2018.

\bibitem{survey_generic_object_detection}
Li Liu, Wanli Ouyang, Xiaogang Wang, Paul Fieguth, Jie Chen, Xinwang Liu, and Matti Pietik{\"a}inen.
\newblock Deep learning for generic object detection: A survey.
\newblock {\em International journal of computer vision}, 128:261--318, 2020.

\bibitem{ground_network}
Georgios Loukas, Stelios Timotheou, and Erol Gelenbe.
\newblock Robotic wireless network connection of civilians for emergency response operations.
\newblock In {\em 2008 23rd International Symposium on Computer and Information Sciences}, pages 1--6, 2008.

\bibitem{lyu2023unmanned}
Mingyang Lyu, Yibo Zhao, Chao Huang, and Hailong Huang.
\newblock Unmanned aerial vehicles for search and rescue: A survey.
\newblock {\em Remote Sensing}, 15(13):3266, 2023.

\bibitem{challenges_murphy}
Thomas Manzini and Robin Murphy.
\newblock Open problems in computer vision for wilderness sar and the search for patricia wu-murad.
\newblock {\em arXiv preprint arXiv:2307.14527}, 2023.

\bibitem{maruvsic2018region}
{\v{Z}}eljko Maru{\v{s}}i{\'c}, Dunja Bo{\v{z}}i{\'c}-{\v{S}}tuli{\'c}, Sven Gotovac, and Ton{\'c}o Maru{\v{s}}i{\'c}.
\newblock Region proposal approach for human detection on aerial imagery.
\newblock In {\em 2018 3rd International Conference on Smart and Sustainable Technologies (SpliTech)}, pages 1--6. IEEE, 2018.

\bibitem{survey_edgeAI_UAVs}
Patrick McEnroe, Shen Wang, and Madhusanka Liyanage.
\newblock A survey on the convergence of edge computing and {AI} for {UAVs}: Opportunities and challenges.
\newblock {\em IEEE Internet of Things Journal}, 9(17):15435--15459, 2022.

\bibitem{attention_object_detection}
Shuyu Miao, Shanshan Du, Rui Feng, Yuejie Zhang, Huayu Li, Tianbi Liu, Lin Zheng, and Weiguo Fan.
\newblock Balanced single-shot object detection using cross-context attention-guided network.
\newblock {\em Pattern recognition}, 122:108258, 2022.

\bibitem{moreira2019performance}
Daniel Moreira, Mateusz Trokielewicz, Adam Czajka, Kevin Bowyer, and Patrick Flynn.
\newblock Performance of humans in iris recognition: The impact of iris condition and annotation-driven verification.
\newblock In {\em 2019 IEEE Winter Conference on Applications of Computer Vision (WACV)}, pages 941--949. IEEE, 2019.

\bibitem{pedestrian_yolo}
Chintakindi~Balaram Murthy, Mohammad~Farukh Hashmi, and Avinash~G Keskar.
\newblock Efficientlitedet: a real-time pedestrian and vehicle detection algorithm.
\newblock {\em Machine Vision and Applications}, 33(3):47, 2022.

\bibitem{driving_yolo}
Jamuna~S Murthy, GM Siddesh, Wen-Cheng Lai, BD Parameshachari, Sujata~N Patil, and KL Hemalatha.
\newblock Objectdetect: A real-time object detection framework for advanced driver assistant systems using yolov5.
\newblock {\em Wireless Communications and Mobile Computing}, 2022, 2022.

\bibitem{nuclear_accident}
Keiji Nagatani, Seiga Kiribayashi, Yoshito Okada, Kazuki Otake, Kazuya Yoshida, Satoshi Tadokoro, Takeshi Nishimura, Tomoaki Yoshida, Eiji Koyanagi, Mineo Fukushima, et~al.
\newblock Emergency response to the nuclear accident at the fukushima daiichi nuclear power plants using mobile rescue robots.
\newblock {\em Journal of Field Robotics}, 30(1):44--63, 2013.

\bibitem{survey_aerial}
Kien Nguyen, Clinton Fookes, Sridha Sridharan, Yingli Tian, Feng Liu, Xiaoming Liu, and Arun Ross.
\newblock The state of aerial surveillance: A survey.
\newblock {\em arXiv preprint arXiv:2201.03080}, 2022.

\bibitem{survey_pedestrian_occlusion}
Chen Ning, Li Menglu, Yuan Hao, Su Xueping, and Li Yunhong.
\newblock Survey of pedestrian detection with occlusion.
\newblock {\em Complex \& Intelligent Systems}, 7:577--587, 2021.

\bibitem{ground_map_reinforcementLearning}
Farzad Niroui, Kaicheng Zhang, Zendai Kashino, and Goldie Nejat.
\newblock Deep reinforcement learning robot for search and rescue applications: Exploration in unknown cluttered environments.
\newblock {\em IEEE Robotics and Automation Letters}, 4(2):610--617, 2019.

\bibitem{humanperception_objectdetection}
Chen Pan and Wei~Qi Yan.
\newblock Object detection based on saturation of visual perception.
\newblock {\em Multimedia Tools and Applications}, 79:19925--19944, 2020.

\bibitem{aerial_groundpenetratingradar}
Maria~Gaia Pensieri, Mauro Garau, and Pier~Matteo Barone.
\newblock Drones as an integral part of remote sensing technologies to help missing people.
\newblock {\em Drones}, 4(2):15, 2020.

\bibitem{piland2023model}
Jacob~C Piland, Adam Czajka, and Christopher Sweet.
\newblock Model focus improves performance of deep learning-based synthetic face detectors.
\newblock {\em IEEE Access}, 11:63430--63441, 2023.

\bibitem{prijatelj2022human}
Derek~S Prijatelj, Samuel Grieggs, Jin Huang, Dawei Du, Ameya Shringi, Christopher Funk, Adam Kaufman, Eric Robertson, and Walter~J Scheirer.
\newblock Human activity recognition in an open world.
\newblock {\em arXiv preprint arXiv:2212.12141}, 2022.

\bibitem{aerial_indoors}
Aveek Purohit, Zheng Sun, Frank Mokaya, and Pei Zhang.
\newblock Sensorfly: Controlled-mobile sensing platform for indoor emergency response applications.
\newblock In {\em Proceedings of the 10th ACM/IEEE International Conference on Information Processing in Sensor Networks}, pages 223--234, 2011.

\bibitem{ground_SMURF}
Frédéric Py, Giulia Robbiani, Giancarlo Marafioti, Yu Ozawa, Masahiro Watanabe, Kenichi Takahashi, and Satoshi Tadokoro.
\newblock Smurf software architecture for low power mobile robots: experience in search and rescue operations.
\newblock In {\em 2022 IEEE International Symposium on Safety, Security, and Rescue Robotics (SSRR)}, pages 264--269, 2022.

\bibitem{redmon2016yolo}
Joseph Redmon, Santosh Divvala, Ross Girshick, and Ali Farhadi.
\newblock You only look once: Unified, real-time object detection.
\newblock In {\em Proceedings of the IEEE conference on computer vision and pattern recognition}, pages 779--788, 2016.

\bibitem{psyphy}
Brandon RichardWebster, Samuel~E. Anthony, and Walter~J. Scheirer.
\newblock Psyphy: A psychophysics driven evaluation framework for visual recognition.
\newblock {\em IEEE Transactions on Pattern Analysis and Machine Intelligence}, 41(9):2280--2286, 2019.

\bibitem{richardwebster2018visual}
Brandon RichardWebster, So~Yon Kwon, Christopher Clarizio, Samuel~E Anthony, and Walter~J Scheirer.
\newblock Visual psychophysics for making face recognition algorithms more explainable.
\newblock In {\em Proceedings of the European conference on computer vision (ECCV)}, pages 252--270, 2018.

\bibitem{russell2024nomad}
Arturo~Miguel Russell~Bernal, Walter Scheirer, and Jane Cleland-Huang.
\newblock Nomad: A natural, occluded, multi-scale aerial dataset, for emergency response scenarios.
\newblock In {\em Proceedings of the IEEE/CVF Winter Conference on Applications of Computer Vision}, pages 8584--8595, 2024.

\bibitem{survey_generic_occlusion}
Kaziwa Saleh, Sándor Szénási, and Zoltán Vámossy.
\newblock Occlusion handling in generic object detection: A review.
\newblock In {\em 2021 IEEE 19th World Symposium on Applied Machine Intelligence and Informatics (SAMI)}, pages 000477--000484, 2021.

\bibitem{deep_cnn_sard}
Sasa Sambolek and Marina Ivasic{-}Kos.
\newblock Automatic person detection in search and rescue operations using deep {CNN} detectors.
\newblock {\em {IEEE} Access}, 9:37905--37922, 2021.

\bibitem{sard}
Sasa Sambolek and Marina Ivasic-Kos.
\newblock Search and rescue image dataset for person detection - {SARD}, 2021.
\newblock {IEEE Dataport}, \url{https://dx.doi.org/10.21227/ahxm-k331}.

\bibitem{walter_original}
Walter~J. Scheirer, Samuel~E. Anthony, Ken Nakayama, and David~D. Cox.
\newblock Perceptual annotation: Measuring human vision to improve computer vision.
\newblock {\em IEEE Transactions on Pattern Analysis and Machine Intelligence}, 36(8):1679--1686, 2014.

\bibitem{survey_civilApplications}
Hazim Shakhatreh, Ahmad~H. Sawalmeh, Ala Al-Fuqaha, Zuochao Dou, Eyad Almaita, Issa Khalil, Noor~Shamsiah Othman, Abdallah Khreishah, and Mohsen Guizani.
\newblock Unmanned aerial vehicles (uavs): A survey on civil applications and key research challenges.
\newblock {\em IEEE Access}, 7:48572--48634, 2019.

\bibitem{aerial_mapping}
Merlin Stampa, Andreas Sutorma, Uwe Jahn, Felix Willich, Sylvia Pratzler-Wanczura, Jörg Thiem, Christof Röhrig, and Carsten Wolff.
\newblock A scenario for a multi-uav mapping and surveillance system in emergency response applications.
\newblock In {\em 2020 IEEE 5th International Symposium on Smart and Wireless Systems within the Conferences on Intelligent Data Acquisition and Advanced Computing Systems (IDAACS-SWS)}, pages 1--6, 2020.

\bibitem{challenges_walter}
Niko S{\"u}nderhauf, Oliver Brock, Walter Scheirer, Raia Hadsell, Dieter Fox, J{\"u}rgen Leitner, Ben Upcroft, Pieter Abbeel, Wolfram Burgard, Michael Milford, et~al.
\newblock The limits and potentials of deep learning for robotics.
\newblock {\em The International journal of robotics research}, 37(4-5):405--420, 2018.

\bibitem{trokielewicz2019perception}
Mateusz Trokielewicz, Adam Czajka, and Piotr Maciejewicz.
\newblock Perception of image features in post-mortem iris recognition: Humans vs machines.
\newblock In {\em 2019 IEEE 10th International Conference on Biometrics Theory, Applications and Systems (BTAS)}, pages 1--8. IEEE, 2019.

\bibitem{yolov7}
Chien-Yao Wang, Alexey Bochkovskiy, and Hong-Yuan~Mark Liao.
\newblock Yolov7: Trainable bag-of-freebies sets new state-of-the-art for real-time object detectors.
\newblock In {\em Proceedings of the IEEE/CVF Conference on Computer Vision and Pattern Recognition (CVPR)}, pages 7464--7475, June 2023.

\bibitem{ground_3dmap}
Hongling Wang, Chengjin Zhang, Yong Song, Bao Pang, and Guangyuan Zhang.
\newblock Three-dimensional reconstruction based on visual slam of mobile robot in search and rescue disaster scenarios.
\newblock {\em Robotica}, 38(2):350--373, 2020.

\bibitem{benefit_usability_tests}
Christian Wankm{\"u}ller, Maximilian Kunovjanek, and Sebastian Mayrg{\"u}ndter.
\newblock Drones in emergency response--evidence from cross-border, multi-disciplinary usability tests.
\newblock {\em International Journal of Disaster Risk Reduction}, 65:102567, 2021.

\bibitem{detectron2}
Yuxin Wu, Alexander Kirillov, Francisco Massa, Wan-Yen Lo, and Ross Girshick.
\newblock Detectron2.
\newblock \url{https://github.com/facebookresearch/detectron2}, 2019.

\bibitem{detectron2modelzoo}
Yuxin Wu, Alexander Kirillov, Francisco Massa, Wan-Yen Lo, and Ross Girshick.
\newblock Detectron2 {ModelZoo}, 2019.
\newblock \url{https://github.com/facebookresearch/detectron2/blob/main/MODEL_ZOO.md}, Last accessed on 2024-09-09.

\bibitem{UCOWOD}
Zhiheng Wu, Yue Lu, Xingyu Chen, Zhengxing Wu, Liwen Kang, and Junzhi Yu.
\newblock Uc-owod: Unknown-classified open world object detection.
\newblock In {\em European Conference on Computer Vision}, pages 193--210. Springer, 2022.

\bibitem{ppyoloe}
Shangliang Xu, Xinxin Wang, Wenyu Lv, Qinyao Chang, Cheng Cui, Kaipeng Deng, Guanzhong Wang, Qingqing Dang, Shengyu Wei, Yuning Du, et~al.
\newblock Pp-yoloe: An evolved version of yolo.
\newblock {\em arXiv preprint arXiv:2203.16250}, 2022.

\bibitem{zhao2024detrs}
Yian Zhao, Wenyu Lv, Shangliang Xu, Jinman Wei, Guanzhong Wang, Qingqing Dang, Yi Liu, and Jie Chen.
\newblock Detrs beat yolos on real-time object detection.
\newblock In {\em Proceedings of the IEEE/CVF Conference on Computer Vision and Pattern Recognition}, pages 16965--16974, 2024.

\bibitem{OSODD}
Jiyang Zheng, Weihao Li, Jie Hong, Lars Petersson, and Nick Barnes.
\newblock Towards open-set object detection and discovery.
\newblock In {\em Proceedings of the IEEE/CVF Conference on Computer Vision and Pattern Recognition (CVPR) Workshops}, pages 3961--3970, June 2022.

\bibitem{zhou2019objects}
Xingyi Zhou, Dequan Wang, and Philipp Kr{\"a}henb{\"u}hl.
\newblock Objects as points.
\newblock {\em arXiv preprint arXiv:1904.07850}, 2019.

\bibitem{occlusionvshumans}
Hongru Zhu, Peng Tang, Jeongho Park, Soojin Park, and Alan Yuille.
\newblock Robustness of object recognition under extreme occlusion in humans and computational models.
\newblock {\em arXiv preprint arXiv:1905.04598}, 2019.

\bibitem{object_detection_survey}
Zhengxia Zou, Keyan Chen, Zhenwei Shi, Yuhong Guo, and Jieping Ye.
\newblock Object detection in 20 years: A survey.
\newblock {\em Proceedings of the IEEE}, 111(3):257--276, 2023.

\end{thebibliography}
